%% file: main_arxiv_v1.tex
\documentclass[10pt]{article} 
\usepackage[preprint]{tmlr}

\usepackage{url}
\usepackage{microtype}
\usepackage{graphicx}
\usepackage{booktabs} 
\usepackage{amsmath}
\usepackage{amssymb}
\usepackage{mathtools}
\usepackage{amsthm}
\usepackage{microtype}
\usepackage{booktabs} 
\usepackage{subcaption}
\usepackage{xcolor}
\usepackage{hyperref}
\usepackage{bbm}
\usepackage{graphicx}
\usepackage{cancel}
\usepackage{wrapfig}
\usepackage{algorithm}
\usepackage{algorithmic}

\input{_util/commands}

\title{Revisiting Critic Estimation in Soft Actor-Critic Algorithms}

\author{\name Jan Felix Kleuker \email j.f.kleuker@liacs.leidenuniv.nl \\
        \name Koen Ponse \email k.ponse@liacs.leidenuniv.nl \\
        \name Aske Plaat \email a.plaat@liacs.leidenuniv.nl \\
        \name Thomas M. Moerland \email t.m.moerland@liacs.leidenuniv.nl \\
        \addr Leiden Institute of Advanced Computer Science (LIACS)\\
        Leiden University}

\definecolor{figtodo}{rgb}{0.00,0.42,0.12}

\newcommand{\SACb}{SAC}
\newcommand{\SACA}{SAC-A}
\newcommand{\SACDA}{SAC-DA}
\newcommand{\SACL}{SAC-$\lambda$}
\newcommand{\SACLR}{SAC-$\lambda$-R}
\newcommand{\SACG}{SAC-GAE}
\newcommand{\SACGR}{SAC-GAE-R}

\begin{document}

\maketitle

\begin{abstract}
Soft Actor-Critic (SAC) is a widely used deep reinforcement learning algorithm that combines entropy regularization with an actor-critic architecture. Standard implementations rely on one-step temporal-difference learning of Q-functions, yet alternative critic design choices remain underexplored. In this work, we revisit three such choices: learning soft advantages instead of soft Q-values, the use of multi-step return targets, and the training of multi-step critics from replay buffer data. Keeping the policy objective fixed, we compare the six critic variants these choices generate: (i) advantage and (ii) dueling-advantage critics learned from one-step transitions, and $\lambda$-return and soft Generalized Advantage Estimation (GAE) critics learned either (iii, iv) on-policy or (v, vi) from replayed trajectory segments. Each variant is evaluated across an entire random hyperparameter search on fourteen tasks, treating the hyperparameters as part of the algorithm. Learning soft advantages instead of soft Q-values does not improve performance, and the soft GAE critic adds nothing over the simpler $\lambda$-return critic. What does improve performance, most clearly on continuous control, is training a $\lambda$-return critic on replayed trajectory segments, even without importance weighting or trace cutting. This extends earlier evidence that uncorrected multi-step targets work well under replay to the full soft $\lambda$-return, entropy terms included. By design, the underlying ideas studied here extend beyond SAC, and our empirical findings aim to encourage their application to a broader class of reinforcement learning algorithms.
\end{abstract}

\section{Introduction}
\label{sec:Intro}

\begin{wrapfigure}[18]{r}{0.4\textwidth}
    \centering
    \vspace{-2.5em}
    \includegraphics[width=\linewidth]{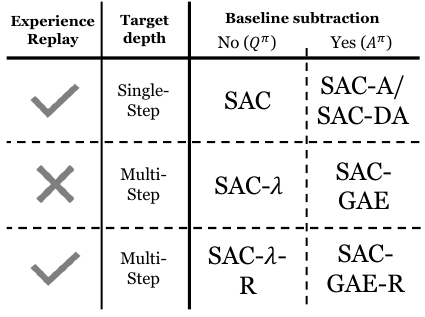}

    \caption{\textbf{Design choices and resulting variants.} The critic is trained on replayed or on-policy data, with one-step or multi-step $\lambda$-return targets, and predicts soft Q-values $Q_\alpha^\pi$ or soft advantages $A_\alpha^\pi$. SAC occupies the one-step, replay-based Q-value cell; the other five cells are the variants we study.}
    \label{fig:00}
\end{wrapfigure}
Soft Actor-Critic (SAC;~\citealt{haarnojaSoft2018}) is one of the most widely adopted algorithms in deep reinforcement learning. Its entropy regularization sustains exploration~\citep{haarnojaSoft2018}, improves robustness~\citep{eysenbachMaxi2022}, and yields a unique optimal policy, unlike the unregularized objective~\citep{jhaveriConv2025}. Modern implementations~\citep{raffin2021stable, bou2023torchrl} follow \citet{haarnojaSoft2019} in learning one or two soft Q-functions from one-step temporal-difference errors. This design is a convention rather than a requirement: SAC is defined by its actor-critic structure and entropy regularization, not by a particular critic. This leaves considerable freedom in how the critic is learned -- freedom that remains largely unexplored.

We study three dimensions of this freedom, which together span the grid of critic variants in Figure~\ref{fig:00}: (i) learning soft advantages instead of soft Q-values; (ii) the rollout depth of the target, through soft $\lambda$-returns; and (iii) whether multi-step targets are computed from on-policy rollouts or replayed from a buffer. 
Since all variants share the policy update, the implementation and the evaluation protocol, differences in outcome can be attributed to the critic.

In A3C~\citep{mnih2016asynchronous} and Proximal Policy Optimization (PPO;~\citealt{schulman2017proximal}), advantages reduce the variance of score-function policy gradients. This role does not carry over to SAC, whose policy gradient is unaffected by state-dependent shifts of the critic (Section~\ref{sec:Background}). Advantages can nevertheless be numerically favorable: SAC's policy depends only on the differences between actions, which can be small relative to the value itself and hard to resolve with an approximate Q-function~\citep{tallec2019making}, whereas an advantage critic learns them directly~\citep{baird1994advantage, wangDuel2016}. We test whether this makes SAC more robust to its hyperparameters, learning soft advantages with an explicit advantage loss (\SACA{}) or jointly with a soft state value in a dueling architecture (\SACDA{}); all other components are inherited from SAC.

The rollout depth of the target trades off bias and variance in the critic's learning signal. Generalized Advantage Estimation (GAE;~\citealt{schulmanHigh2018}) interpolates between one-step bootstrapped targets and Monte Carlo returns, and PQN~\citep{gallici2024simplifying} recently carried the same idea over to Q-learning, replacing the replay buffer with many parallel environments. Following the idea of PQN, we drop the replay buffer, and train a soft Q-function on soft $\lambda$-returns from parallel on-policy rollouts (\SACL{}). Combining both axes, we also implement a variant in which we estimate soft advantages using GAE (\SACG{}).

Multi-step targets are harder to combine with replay, since a replayed trajectory was generated by an older policy and biases the target. Many methods correct for this with importance weights or truncated traces~\citep{espeholt2018impala}, including concurrent work on SAC~\citep{lyskawaSACn2025}, yet uncorrected multi-step targets from a buffer have been demonstrated to work well in value-based methods~\citep{hessel2018rainbow, fedus2020revisiting, clark2024beyond} and in actor-critic continuous control~\citep{kozuno2021revisiting}. For the entropy-regularized target, the picture is less settled: \citet{kozuno2021revisiting} found its per-step entropy terms destabilizing and kept only the final one. We train both multi-step critics on replayed segments with the full soft $\lambda$-return and no correction (\SACLR{}, \SACGR{}), and compare them with their on-policy counterparts to see whether the bias outweighs the benefit of reusing data.

All of these choices solely concern critic learning and fit into existing SAC implementations with little overhead. We present them for reinforcement learning with a general convex regularizer, of which SAC's entropy bonus is the most prominent instance, and then measure what each buys in SAC across an entire random hyperparameter search on fourteen tasks from three benchmark suites. Soft advantages improve on soft Q-values neither in return nor in robustness, and the GAE critic adds nothing over the $\lambda$-return critic. In continuous control, multi-step gains come with replay rather than depth: within our sampled ranges, neither a larger $\lambda$ nor a longer horizon helps, whereas training on replayed segments does, despite the uncorrected bias. This extends earlier evidence for uncorrected multi-step targets under replay to the full soft $\lambda$-return. In practice, the critic parameterization can be chosen on grounds other than return, and in SAC, too, multi-step targets can be paired with replay without an off-policy correction.

\section{Preliminaries}
\label{sec:Background}

We assume a standard Markov Decision Process (MDP) defined by a tuple $\mathcal{M} = (\S,\A,p_0,p,\gamma)$. Here, $\S$ is the state space, $\A$ is the action space, $p_0 \in \Delta(\S)$ 
is the initial state distribution, $p(\cdot,\cdot|s,a) \in \Delta(\S\times\R)$ is the probabilistic transition and reward function, and $\gamma \in [0,1)$ is the discount factor. We usually abbreviate the reward by $r(s,a) = \IE_{s',r\sim p(s',r|s,a)}[r]$. 

The behavior of an agent interacting with this MDP is defined by a policy, that is, a mapping $\pi: \S \to \Delta(\A)$ assigning a distribution over actions to each state. Each policy induces a distribution $p_\pi(\tau)$ on the set of trajectories $\tau = (s_0,a_0,s_1,\dots)$ \citep{agarwalRein}. The value function is defined as $V^\pi(s) = \IE_{\tau \sim p_\pi(\tau)|s_0=s}[\sum_{t=0}^\infty \gamma^t \,r(s_t,a_t)]$, resulting in the RL objective $\pi^\ast(\cdot|s) = \arg\max_{\pi \in \Pi} V^\pi(s)$.

\textbf{Regularized MDP.} A natural extension of the RL objective adds a regularization term at the reward level, also referred to as MDP regularization \citep{ziebart2010modeling, leeTsal2019}. 
More precisely, given a convex function $h: \Delta(\A) \to \R$, the corresponding regularized value function for some regularization weight (temperature) $\alpha \geq 0$ is defined as $V_\alpha^\pi(s) := \IE_{\tau\sim p_\pi(\tau)| s_0=s}[\sum_{t=0}^\infty \gamma^t \{r(s_t,a_t) - \alpha\,h(\pi(\cdot|s_t))\} ]$ \citep{neu2017unified, lanPoli2023}. The resulting regularized MDP objective can be expressed as $\pi^\ast(\cdot|s) = \arg\max_{\pi\in\Pi} V_\alpha^\pi(s)$. 

Following the convention of prior literature \citep{haarnojaSoft2018, meiGlob2022, zhan2023policy}, we define the regularized Q-function as $Q_\alpha^\pi(s,a) := r(s,a) + \gamma \IE_{s'\sim p(\cdot|s,a)}[V_\alpha^\pi(s')]$. This entails the relationship $V_\alpha^\pi(s) = \IE_{a\sim\pi(\cdot|s)}[Q_\alpha^\pi(s,a) - \alpha h(\pi(\cdot|s))]$ between the two soft functions (e.g. \citealt{zhan2023policy}), and gives rise to the soft Bellman Operator \citep{haarnojaSoft2018}:
\begin{equation}
    \label{eq:2}
    \begin{aligned}
        T_\alpha^\pi Q(s,a) = r(s,a) + \gamma \IE_{s' \sim p(\cdot|s,a), a'\sim\pi(\cdot|s')}[Q(s',a') - \alpha h(\pi(\cdot|s'))].
    \end{aligned}
\end{equation}

Throughout, we fix $h$ to the negative Shannon entropy, $h(p) = \IE_{x \sim p}[\log(p(x))] = - \mathcal{H}(p)$ for $p \in \Delta(\A)$, which biases the search towards more exploratory policies and forms the basis of the SAC algorithm \citep{ziebart2010modeling, haarnojaRein2017, haarnojaSoft2018}.

\subsection{Soft action-values and soft advantages}

Advantages measure how much better an action is than the policy's average in a given state, separating the action-dependent part of the value from its state-dependent level. For soft Q-functions, the difference $Q_\alpha^\pi - V_\alpha^\pi$ does not generally vanish in expectation, so we follow prior work~\citep{meiGlob2022, choeMaxi} and define the soft advantages as
\begin{equation}
    A^\pi_\alpha(s,a) := Q^\pi_\alpha(s,a) - \IE_{a\sim \pi(\cdot|s)}[Q^\pi_\alpha(s,a)].
\end{equation}
For $\alpha = 0$ this recovers the standard advantage, since then $V^\pi(s) = \IE_{a\sim \pi(\cdot|s)}[Q^\pi(s,a)]$. As $Q^\pi_\alpha$ and $A^\pi_\alpha$ differ only by a state-dependent term, soft advantages can replace the soft Q-function in the SAC policy update:

\begin{equation}
    \label{eq:4}
    \begin{aligned}
        \pi_{k+1}(\cdot|s) &= \arg\min_{\pi\in\Pi} \IE_{a \sim \pi(\cdot|s)}[-Q_\alpha^{\pi_k}(s,a)] + \alpha \,h(\pi(\cdot|s))\\
        &= \arg\min_{\pi\in\Pi} \IE_{a \sim \pi(\cdot|s)}[-A_\alpha^{\pi_k}(s,a)] + \alpha \,h(\pi(\cdot|s)).
    \end{aligned}
\end{equation}
The two forms yield identical gradients, exactly for discrete actions and for every sample of SAC's reparameterized estimator, so a Q-critic and an advantage critic can differ only in how well they are learned. We will use the following properties of soft advantages.

\begin{lemma}
    \label{lem:2.1}
    The soft advantages $A^\pi_\alpha$ satisfy the following properties for any $(s,a)\in\S\times\A$, $\alpha \geq 0$:
    \begin{enumerate}
        \item[(i)] The soft advantages can be obtained from the soft value function as $A^\pi_\alpha(s,a) = r(s,a) - \alpha h(\pi(\cdot|s)) + \gamma \IE_{s'\sim p(\cdot|s,a)}[V_\alpha^\pi(s')] - V^\pi_\alpha(s)$,
        \item[(ii)] The soft advantages vanish in expectation, i.e. $\IE_{a\sim\pi(\cdot|s)} [A^\pi_\alpha(s,a)] = 0$.
    \end{enumerate}
\end{lemma}

Property (i) expresses the soft advantages in terms of the soft value function and is the key ingredient for several algorithms in the next section. Property (ii) confirms that they retain only the action-dependent part of the Q-function.
For completeness, we provide proofs in Appendix~\ref{sec:appendix-proofs}.

\begin{figure*}[tbp]
    \centering
    \includegraphics[width=0.95\linewidth]{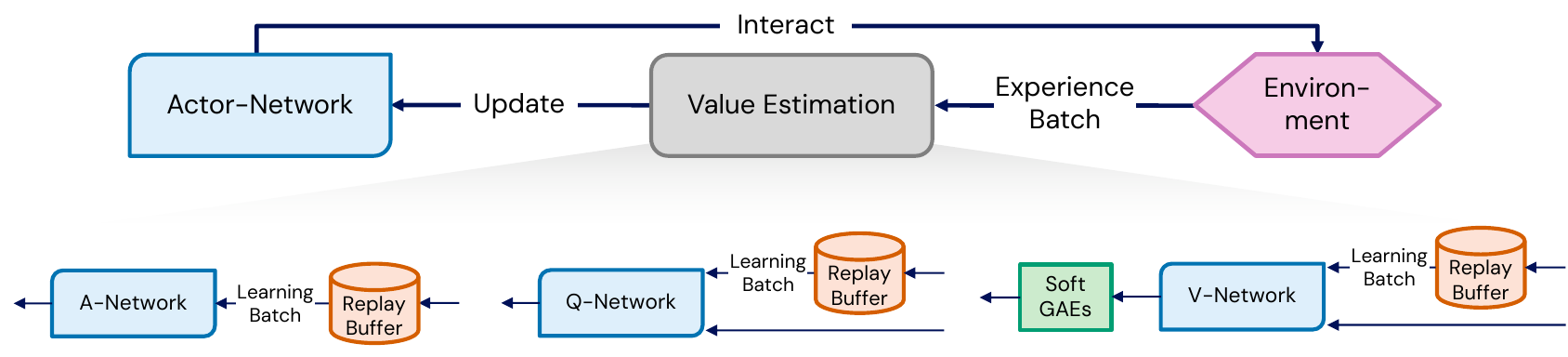}
    \hspace{0.15\linewidth}\textbf{(a)} \SACA{} \& \SACDA{} \hspace{0.04\linewidth}\textbf{(b)} \SACb{}, \SACL{} \& \SACLR{} \hspace{0.05\linewidth} \textbf{(c)} \SACG{} \& \SACGR{}
    \caption{\textbf{Training pipelines of the SAC variants.} (a) \SACA{} and \SACDA{} keep SAC's replay-based one-step pipeline but learn soft advantages. (b) SAC trains its Q-network on replayed transitions, \SACL{} on parallel on-policy rollouts, and \SACLR{} on replayed trajectory segments. (c) \SACG{} and \SACGR{} learn a soft value network, on rollouts or replayed segments, and compute soft GAEs for the policy update.}
    \label{fig:OverviewAlgs}
\end{figure*}

\section{Estimating the targets}
\label{sec:04}

The soft Bellman operator of Eq.~\ref{eq:2} and Lemma~\ref{lem:2.1} are all that is needed to train critics other than SAC's one-step Q-function.  We now formalize the variants of Figure~\ref{fig:00}, with training pipelines shown in Figure~\ref{fig:OverviewAlgs}: after the SAC critic, we introduce one-step advantage critics, then critics trained on soft $\lambda$-returns and soft GAE, and finally their replay-based counterparts. All variants share the policy objective of Eq.~\ref{eq:4}; pseudo-code and the remaining implementation differences are given in Appendix~\ref{sec:appendix-algdetails}.

\subsection{Soft Actor-Critic} 

Following \citet{haarnojaSoft2019}, in the now common form without a separate value network, SAC trains $Q_\theta$ on the soft one-step temporal difference error of Eq. \ref{eq:2}
\begin{equation}
    \label{eq::SACLoss}
    \begin{aligned}
        L^\mathrm{SAC}_\mathrm{critic}(\theta) = \IE_{(s,a,s')\sim\mathcal{D}}\Bigl[\Bigl\{Q_\theta(s,a) - r(s,a)- \gamma \IE_{a'\sim\pi(\cdot|s')}\bigl[ Q_{\bar{\theta}}(s',a') \bigr] +\alpha\gamma h\bigl(\pi(\cdot|s')\bigr)\Bigr\}^2\Bigr].
    \end{aligned}
\end{equation}
Here, $\bar{\theta}$ refers to a target network, and $\mathcal{D}$ to a buffer of stored one-step transitions \citep{raffin2021stable, bou2023torchrl}. Throughout, the expectations over $\pi$ are estimated by sampling for continuous actions.

\subsection{Soft Actor-Critic with Advantages}
\label{sec:advantages}

Based on Lemma \ref{lem:2.1} we can provide a Bellman-like equation for the soft values, and hence a loss for an approximation of the soft values $V_\phi$:
\begin{equation}
    \label{eq::LV}
    \begin{aligned}
        L_\mathrm{V}(\phi) = \IE_{(s,a,s')\sim \mathcal{D}}\Bigr[ \{V_\phi(s) - r(s,a) + \alpha h(\pi(\cdot|s)) - \gamma V_{\bar{\phi}}(s') \}^2\Bigl].
    \end{aligned}
\end{equation}
Using the value estimates, an advantage network $A_\theta$ can be trained, based on Lemma \ref{lem:2.1}:
\begin{equation}
    \label{eq:LAalpha}
    L^\mathrm{A}_\mathrm{critic}(\theta) = \IE_{(s,a,s')\sim \mathcal{D}}\Bigl[\bigl\{A_{\theta}(s,a) - r(s,a) + \alpha h(\pi(\cdot|s)) - \gamma V_\phi(s') + V_\phi(s)\bigr\}^2\Bigr].
\end{equation}
Since Eq.~\ref{eq::LV} uses the action stored in the buffer, under replay $V_\phi$ targets the soft value of the behavior policy rather than of $\pi$, and \SACA{} inherits this bias. The dueling formulation below avoids it, as its combined critic is trained on the standard soft Q-target, which is valid off-policy.

\textbf{Dueling Q-Networks.} Advantage and value networks can also be trained under a single objective, following the dueling architecture of \citet{wangDuel2016}, in which the two are recombined into a Q-function that is then trained on the one-step temporal difference error. 
Since $Q^\pi_\alpha(s,a) = A^\pi_\alpha(s,a) + V^\pi_\alpha(s) + \alpha h(\pi(\cdot|s))$ by definition, we set
\begin{equation}
    \label{eq::DAJointQ}
    \begin{aligned}
        Q_{\theta,\phi}(s,a) = A_\theta(s,a) + V_\phi(s) + \alpha h(\pi(\cdot|s))  - \kappa_\mathrm{d}\,\widehat{\IE}\bigl[A_\theta(s,\cdot)\bigr],
    \end{aligned}
\end{equation}
where the last term is the regularizer of \citet{wangDuel2016}. 
For a finite action set we compute it exactly, $\widehat{\IE}[A_\theta(s,\cdot)] = \frac{1}{|\A|}\sum_{a' \in \A}A_\theta(s,a')$; for continuous actions the mean is estimated from samples $a' \sim \pi(\cdot|s)$. The combined critic is trained with the SAC loss of Eq.~\ref{eq::SACLoss},
\begin{equation}
    \label{eq::DASACLoss}
    \begin{aligned}
    L^\mathrm{DA}_\mathrm{critic}(\theta,\phi) &= \IE_{(s,a,s')}\Bigl[\Bigl\{Q_{\theta,\phi}(s,a) - r(s,a) - \gamma \IE_{a'\sim\pi}\bigl[ Q_{\bar{\theta}, \bar{\phi}}(s',a') \bigr] +\alpha \gamma h\bigl(\pi(\cdot|s')\bigr)\Bigr\}^2\Bigr].
    \end{aligned}
\end{equation}
In our experiments, $\kappa_\mathrm{d} = 0$ worked best, leaving the split into $A_\theta$ and $V_\phi$ unidentifiable: \SACDA{} thus tests the dueling parameterization of the soft Q-critic, while \SACA{} learns soft advantages explicitly. By design, both variants use a single advantage and value network instead of the two SAC critics; otherwise, they train on one-step transitions from a replay buffer and drop into existing SAC implementations unchanged (Figure~\ref{fig:OverviewAlgs}a). We refer to them as \SACA{} and \SACDA{}.

\subsection{Soft Actor-Critic with \texorpdfstring{$\lambda$}{lambda}-returns}

The second dimension is the rollout depth of the target (Figure~\ref{fig:00}, middle row), for which we adopt the two changes that lead from DQN to PQN. First, we replace the one-step target with a truncated $\lambda$-return~\citep{sutton1988learning, watkins1989learning, seijen2014true}, defined through the soft Bellman operator as
\begin{equation}
    \label{eq:lambdareturnoperator}
    T_{\lambda,\alpha}^\pi Q := (1-\lambda) \sum_{n=1}^{N-1}\lambda^{n-1} (T_\alpha^\pi)^{n} Q + \lambda^{N-1} (T_\alpha^\pi)^{N} Q ,
\end{equation}
where $\lambda \in [0,1]$, $N$ is the horizon, and $(T_\alpha^\pi)^n$ applies Eq.~\ref{eq:2} $n$ times. Second, we replace the replay buffer by parallel environments (Figure~\ref{fig:OverviewAlgs}b).

For our choice $h=-\mathcal{H}$, the regularizer contributes the entropy bonus $\alpha\mathcal{H}(\pi(\cdot|s))$ at every visited state, and a trajectory sampled from $\pi$ yields an unbiased estimate of $(T_\alpha^\pi)^n Q(s_t,a_t)$, the soft $n$-step return
\begin{equation}
    \label{eq::n-step Returns}
    R_{\alpha,t}^{(n)} := \sum_{k=0}^{n-1}\gamma^k \bigl\{r_{t+k} + \gamma\alpha\mathcal{H}\bigl(\pi(\cdot|s_{t+k+1})\bigr)\bigr\} + \gamma^n \IE_{a\sim\pi(\cdot|s_{t+n})}\bigl[Q(s_{t+n}, a)\bigr].
\end{equation}
The soft $\lambda$-return $R_{\alpha,t}^\lambda := (1-\lambda) \sum_{n=1}^{N-1}\lambda^{n-1} R_{\alpha,t}^{(n)} + \lambda^{N-1} R_{\alpha,t}^{(N)}$ then satisfies the recursion
\begin{equation}
    \label{eq::n-step Returns recursive}
    R_{\alpha,t}^\lambda = r_t + \gamma \bigl\{(1-\lambda)\IE_{a\sim\pi(\cdot|s_{t+1})}[Q(s_{t+1},a)] + \alpha\mathcal{H}\bigl(\pi(\cdot|s_{t+1})\bigr) + \lambda R_{\alpha,t+1}^\lambda\bigr\}
\end{equation}
with $R_{\alpha,t}^\lambda = r_t$ if $s_{t+1}$ is terminal, which allows it to be computed efficiently backwards along a trajectory (Appendix~\ref{sec:appendix-proofs}). This is the soft analogue of Peng's Q($\lambda$)~\citep{peng1994incremental}, the target that PQN uses with a maximum and that \citet{kozuno2021revisiting} combine with SAC. The critic is trained on
\begin{equation}
    \label{eq::LossPSAC}
    L_\mathrm{critic}^{\lambda}(\theta) = \IE_{\tau}\Bigr[\sum_{t=0}^{|\tau|} \{Q_\theta(s_t,a_t) - R_{\alpha,t}^\lambda\}^2\Bigl],
\end{equation}
with targets computed by target networks (Appendix~\ref{sec:appendix-algdetails}); we refer to this variant as \SACL{}.

\subsection{Generalized Soft Advantage Estimation}

Combining multi-step targets with soft advantages (Figure~\ref{fig:00}, middle right) allows the advantages themselves to be estimated by Generalized Advantage Estimation (GAE;~\citealt{schulmanHigh2018}). Formally, for $\lambda \in (0,1]$ given a trajectory $\tau = (s_1,a_1,\dots,s_T)$ we estimate the advantages via
\begin{equation}
    A^\lambda(s_t,a_t) = \sum_{i=0}^{T-t-1} (\lambda\,\gamma)^i \delta(s_{t+i},a_{t+i}),
\end{equation}
where $\delta(s_t,a_t) = r(s_t,a_t) + \gamma V(s_{t+1}) - V(s_t)$.
For $\lambda=1$ it recovers the $(T-t)$-step advantage estimate of \citet{mnih2016asynchronous}, $A^{\lambda=1}(s_t,a_t)= Q^{T-t}(s_t,a_t) - V(s_t)$.
Following \citet{choeMaxi} and property (i) of Lemma~\ref{lem:2.1}, this extends to soft advantages as
\begin{equation}
    \label{eq:SoftGAE}
    \begin{aligned}
        A^\lambda_\alpha(s_t,a_t) &:= \sum_{i=0}^{T-t-1} (\lambda\,\gamma)^i \delta_\alpha(s_{t+i},a_{t+i}),
    \end{aligned}
\end{equation}
where $\delta_\alpha(s_t,a_t) := r(s_t,a_t) - \alpha h(\pi(\cdot|s_t)) + \gamma V^\pi_\alpha(s_{t+1}) - V^\pi_\alpha(s_t)$; again, $\lambda=1$ recovers the analogous $(T-t)$-step expression (Appendix~\ref{sec:appendix-proofs}).
We train a soft value network on the soft $\lambda$-returns, the multi-step counterpart of Eq.~\ref{eq::LV}, use it to compute soft GAEs from sampled rollouts (Figure~\ref{fig:OverviewAlgs}c) and refer to this variant as \SACG{}. Since GAE provides advantages only at sampled actions, \SACG{} updates its policy with the soft score-function gradient~\citep{shiSoft2019, choeMaxi} (Appendix~\ref{sec:appendix-algdetails}); here, unlike in \SACA{} and \SACDA{}, the advantage acts as a variance-reducing baseline.

\subsection{Experience Replay}
\label{sec:replay}

We consider \SACb{} an off-policy algorithm, because its one-step target can be computed from transitions generated by any policy: the current policy only enters at the next state, through the expectation over $a' \sim \pi(\cdot|s')$. \SACDA{} inherits this property; \SACA{} only in part (Section~\ref{sec:advantages}). A multi-step target, in contrast, follows the actions taken along a trajectory, so without a correction such as importance weighting or trace cutting~\citep{espeholt2018impala, lyskawaSACn2025}, it is unbiased only for trajectories sampled from the policy being evaluated.

Replay is nonetheless attractive, since reusing data typically improves sample efficiency. Following other methods that combine uncorrected multi-step targets with a replay buffer~\citep{hessel2018rainbow, fedus2020revisiting, kozuno2021revisiting, clark2024beyond}, we train both multi-step critics on trajectory segments drawn from a buffer and deliberately apply no correction. The result is an off-policy estimate of an on-policy target, which is biased by the age of the stored segments; for \SACGR{}, whose score-function update also uses the replayed actions, the same holds for the policy gradient. Whether that bias matters in practice is an empirical question we return to in the next section. These two variants, \SACLR{} and \SACGR{}, complete the grid in Figure~\ref{fig:00}.

\section{Experiments}
\label{sec:Experiments}

We evaluate all variants on fourteen tasks from three suites: classic control (Acrobot, CartPole, MountainCar, Pendulum;~\citealt{osband2020bsuite}) and MinAtar (Freeway, Asterix, Breakout, SpaceInvaders;~\citealt{young19minatar}), both from Gymnax~\citep{gymnax2022github}, and six MuJoCo control tasks from Brax (Ant, HalfCheetah, Hopper, Humanoid, Reacher, Walker2d;~\citealt{brax2021github}). 
Since Pendulum and Brax tasks have continuous action spaces and the remaining seven tasks discrete ones, we report the two groups separately. As external reference points we include PPO and, on discrete tasks, PQN, sampled under the same protocol. All implementations build on Jaxnasium~\citep{Ponse_Jaxnasium_Utilities_for}; additional results, experimental details, and pseudo-code are given in Appendices~\ref{appendix:additional-plots}, \ref{appendix:experiment-details} and~\ref{sec:appendix-algdetails}.

\textbf{Evaluation protocol.}
We treat hyperparameters as part of the algorithm rather than tuning them away before reporting: a method that performs well only in a narrow region of its hyperparameter space demands more expertise to deploy on a new task than one that performs well broadly. For a fixed algorithm, let $X_{h,m}$ denote the normalized return after training with hyperparameters $h\in\Lambda$ on task $m\in\mathcal{M}$. Even for fixed $h$ and $m$ it is random, through initialization, training seeds and hardware nondeterminism~\citep{agarwalDeep2021}, and through the stochasticity of environment and policy at evaluation. We therefore treat $X_{h,m}:\Omega_\mathrm{train}\times\Omega_\mathrm{test}\to\mathbb{R}$ as a random variable and report properties of its distribution over $h$ and $m$. Returns are normalized per task so that zero corresponds to the lowest achievable return and one to the $95$th percentile of per-run mean returns pooled over all algorithms, which ties the scale neither to a single method nor to a single noisy draw.

Configurations are drawn by Sobol sampling from a region centered on a configuration tuned once for \SACb{}, with no prior tuning of any variant, which amounts to roughly $27{,}000$ training runs; all sampled ranges are listed in Appendix~\ref{appendix:experiment-details}. Two properties of this protocol shape how the results should be read: Any statement about a distribution over hyperparameters is conditional on $\Lambda$: widening it lowers every score, and re-centering it can reorder methods. Centering on the baseline's optimum is the conservative choice with respect to our variants. Moreover, $\Lambda$ does not have the same dimension for every method: variants with a second critic carry additional learning rates and update counts, and are therefore sampled more sparsely at a fixed budget of $H$ configurations per task, which again works against them. Each metric is defined in the subsection that reports it.

\subsection{Robustness to Hyperparameter Selection}

We first ask how likely a randomly drawn configuration is to work at all. The tail probability $\mathbb{P}_{(h,m)\sim U(\Lambda,\mathcal{M})}(X_{h,m} > \tau)$ is the probability that an algorithm reaches performance level $\tau$ on a randomly drawn task under uniformly drawn hyperparameters. We estimate it with an extended version of the score distribution of \citet{agarwalDeep2021},
\begin{equation}
    \label{eq:score-distr}
    \hat{F}_X(\tau) := \frac{1}{M}\sum_{m=1}^{M}\frac{1}{H }\sum_{h=1}^{H}\frac{1}{S}\sum_{s=1}^{S} \frac{1}{S'}\sum_{s'=1}^{S'} \mathbbm{1}_{\{x_{h,m,s,s'} > \tau\}},
\end{equation}
where $x_{h,m,s,s'}$ is the normalized return of configuration $h$ on task $m$ under training seed $s$ and evaluation seed $s'$, and $M$, $H$, $S$, and $S'$ denote the numbers of tasks, configurations per task, training seeds per configuration, and post-training evaluations per training seed.

\begin{figure}[t]
  \centering
  \includegraphics[width=0.95\textwidth]{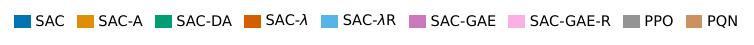}
  \\[0.3em]
  \begin{subfigure}{0.24\textwidth}\includegraphics[width=\linewidth]{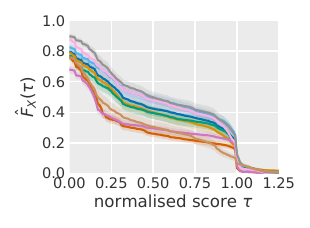}\caption{Discrete tasks}\end{subfigure}\hfill
  \begin{subfigure}{0.24\textwidth}\includegraphics[width=\linewidth]{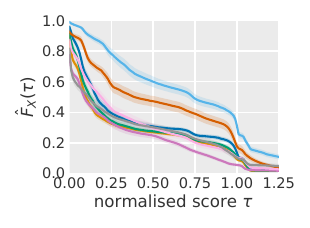}\caption{Continuous tasks}\end{subfigure}\hfill
  \begin{subfigure}{0.24\textwidth}\includegraphics[width=\linewidth]{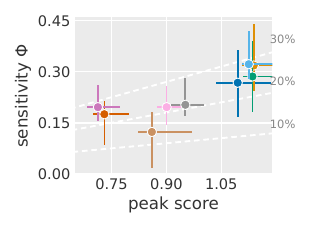}\caption{Discrete tasks}\end{subfigure}\hfill
  \begin{subfigure}{0.24\textwidth}\includegraphics[width=\linewidth]{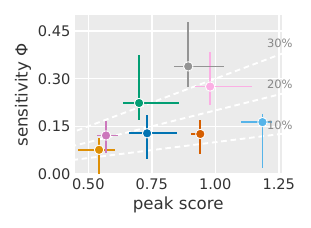}\caption{Continuous tasks}\end{subfigure}
  \caption{\textbf{Performance profiles and peak performance.} (a) \& (b) Performance profiles: the share of configurations reaching normalized score $\tau$ (Eq.~\ref{eq:score-distr}). (c) \& (d) Peak score against sensitivity $\Phi$; dashed lines mark constant relative sensitivity $\Phi/\mathrm{Peak}$, and the lower right corner is best. 
  Bands and bars are 95\% stratified bootstrap confidence intervals. PQN is only defined for discrete action spaces and is omitted for continuous action spaces.
  }
    \label{fig-exp:1}
\end{figure}

In discrete-action tasks (Fig.~\ref{fig-exp:1}a), no SAC variant rises clearly above \SACb{}, while the two on-policy multi-step variants, \SACL{} and \SACG{}, fall clearly below it. On continuous-action tasks (Fig.~\ref{fig-exp:1}b), \SACLR{} dominates every other method at every threshold and \SACL{} is second, both holding a larger share of workable configurations than \SACb{}, the advantage variants and PPO. Two contrasts within the figure already localize the effect. Exchanging soft Q-values for soft advantages moves nothing: \SACA{} and \SACDA{} stay close to \SACb{} throughout. Adding a replay buffer moves a great deal: \SACLR{} and \SACGR{} sit above \SACL{} and \SACG{} across the whole range, without any importance weighting or trace cutting of the replayed targets.

\subsection{Peak vs.\ Sensitivity}

A high chance of drawing a well-performing configuration does not need to come with a high ceiling. Peak performance measures the best score attainable per task,
\begin{equation}
    \label{eq::TS}
    \mathrm{Peak}_m(X) = \max_h x_{h,m} \qquad \mathrm{Peak}(X) = \frac{1}{M} \sum_{m=1}^M \mathrm{Peak}_m(X),
\end{equation}
with $x_{h,m} = \frac{1}{S}\sum_{s=1}^S\frac{1}{S'}\sum_{s'=1}^{S'}x_{h,m,s,s'}$ the empirical mean performance of a configuration on a task. Since peak performance permits per-task tuning, it says little about how well a configuration generalizes across tasks. We therefore pair it with the sensitivity \citep{adkins2024method}, the performance lost when a single configuration must serve all tasks,
\begin{equation}
    \label{eq::Phi}
    \Phi(X) := \mathrm{Peak}(X) - \max_h \frac{1}{M}\sum_{m=1}^M x_{h,m}.
\end{equation}

In discrete tasks (Fig.~\ref{fig-exp:1}c), peaks cluster near $1.1$ for \SACb{}, \SACA{}, \SACDA{} and \SACLR{}, with relative sensitivities of roughly $20$--$30\%$. The on-policy multi-step variants lose about a third of that peak, and the GAE variants trade peak for a modest reduction in sensitivity without reaching PPO.

In continuous tasks (Fig.~\ref{fig-exp:1}d) the picture is clearer. \SACLR{} reaches a clearly higher peak than \SACb{} at roughly the same absolute sensitivity. Because $\Phi$ is an absolute quantity, it flatters methods whose ceiling is low to begin with, so we read it relative to the peak: here \SACLR{} leads with $0.138$, against $0.175$ for \SACb{} and $0.379$ for PPO, and is thus at once the strongest variant and among the cheapest to tune. It ranks first in peak score on five of the seven continuous tasks (Fig.~\ref{fig:App03}) and leads at every random-search budget (Fig.~\ref{fig:App04}). \SACGR{} attains the second-highest peak, far above \SACG{}, though at a higher sensitivity; replacing on-policy rollouts by replayed segments thus lifts both target types. The advantage variants do not improve on \SACb{} here either: \SACA{}, which by design has a single critic, reaches a lower peak, and \SACDA{} a similar peak at a higher sensitivity.

\begin{figure}[t]
  \centering
  \includegraphics[width=0.95\textwidth]{images_v2/legend_discrete.pdf}
  \\[0.3em]
  \begin{subfigure}{0.24\textwidth}\includegraphics[width=\linewidth]{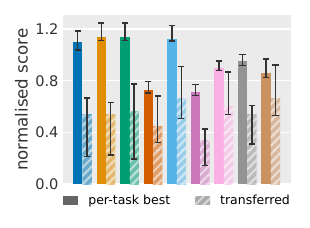}\caption{Discrete Tasks}\end{subfigure}\hfill
  \begin{subfigure}{0.24\textwidth}\includegraphics[width=\linewidth]{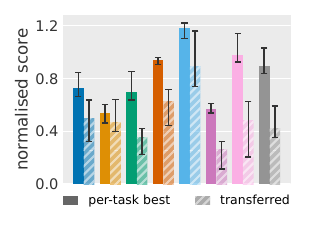}\caption{Continuous Tasks}\end{subfigure}\hfill
  \begin{subfigure}{0.24\textwidth}\includegraphics[width=\linewidth]{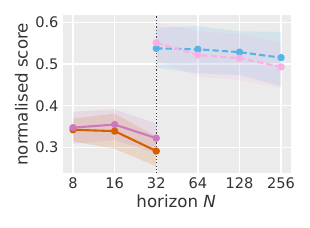}\caption{Discrete Tasks}\end{subfigure}\hfill
  \begin{subfigure}{0.24\textwidth}\includegraphics[width=\linewidth]{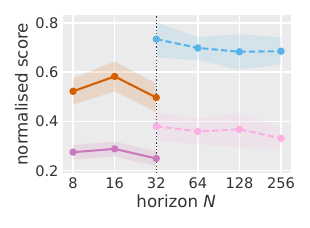}\caption{Continuous Tasks}\end{subfigure}
  \caption{\textbf{Hyperparameter transfer performance and effect of the horizon.} (a) \& (b) Best per-task configuration against the best configuration transferred from all other tasks. (c) \& (d) Normalized score against target horizon $N$. The on-policy variants (solid) are sampled at $N\leq32$, the replay-based ones (dashed) at $N\geq32$. Bands and bars are 95\% stratified bootstrap confidence intervals.}
    \label{fig-exp:2}
\end{figure}

\subsection{Hyperparameter Transfer Performance}
\label{sec:transfer}

Sensitivity charges one configuration with serving every task at once. A milder question is what a configuration tuned on other tasks is worth on an unseen one. Let $h_m^- = \arg\max_h \frac{1}{M-1} \sum_{m'\neq m} x_{h,m'}$ be the best shared configuration over all tasks except $m$; its score on $m$ defines the transfer performance
\begin{equation}
    \label{eq:TP}
    T_m(X) = x_{h_m^-, m}, \qquad T(X) = \frac{1}{M} \sum_{m=1}^M T_m(X).
\end{equation}

Figure~\ref{fig-exp:2}a plots transfer against peak score. On discrete tasks, \SACA{} and \SACDA{} match \SACb{} on both bars, and the replay-based variants retain a larger share of their peak than the one-step variants, even where that peak is lower, as for \SACGR{}; given the intervals, however, no variant separates from \SACb{}. On continuous tasks (Fig.~\ref{fig-exp:2}b), \SACLR{} leads on both bars and retains about $75\%$ of its peak after transfer, followed by \SACL{}, whereas PPO loses more than half of its peak.

\subsection{A closer look at the multi-step variants}
\label{sec:mechanism}

In continuous tasks, the $\lambda$-return critics, and \SACLR{} in particular, are the clearest improvement over \SACb{}. Three ingredients set them apart: the $\lambda$-weighting of the target, its horizon, and, for \SACLR{} and \SACGR{}, the replay buffer. All three vary within our search (Appendix~\ref{appendix:experiment-details}), so each can be examined without further runs, within the sampled ranges.

\textbf{More weight on long returns does not help.} Since $\lambda$ is sampled independently of every other hyperparameter, its effect can be read off directly. Performance is flat or mildly decreasing in $\lambda$ on both splits: of the sixteen slope estimates in Appendix~\ref{app:sec:lambdaablation}, two exclude zero, and both are negative.

\textbf{Nor does a longer horizon.} The on-policy variants use rollouts of $8$ to $32$ steps and the replay-based ones replay segments of $32$ to $256$, so an aggregate comparison between the two families conflates replay with a horizon up to eight times longer. Within each family, the score does not show a consistent increase with the horizon $N$ on either split (Fig.~\ref{fig-exp:2}c,d).

\textbf{Replay does, despite the uncorrected bias.} At $N=32$, the one horizon both families share, the replay-based variants lead for both the $\lambda$-return and the GAE critic and on both splits (Fig.~\ref{fig-exp:2}c,d), so the gain is not due to the horizon. Across all fourteen tasks, a randomly drawn run of a replay-based variant outperforms one of its on-policy counterparts for both critics (see also Fig.~\ref{fig:App02}). This gain persists despite the uncorrected bias of the replayed targets. Since the replay-based variants also use larger batches and their own update schedule, this comparison bounds the combined effect of replay rather than isolating the bias.

These results neither make the multi-step target dispensable nor isolate its contribution. Both the $\lambda$ and the horizon ranges begin well above the one-step limit, so they show only that additional depth beyond them buys nothing. Relative to the one-step \SACb{}, \SACLR{} is clearly ahead on continuous tasks (Fig.~\ref{fig-exp:1}b,d), but the two differ in more than their target, since the replay-based $\lambda$-return critic is trained on sampled segments, with larger batches and its own update schedule. We therefore attribute the gain of \SACLR{} on continuous control to the combination it implements, $\lambda$-return targets computed on replayed segments, rather than to any single ingredient. What our data does rule out is that this gain depends on a large $\lambda$, a long horizon, or an off-policy correction.

\section{Related Work}
\label{sec:RelatedWork}

\textbf{SAC as an instance of Policy Mirror Descent.} Policy Mirror Descent (PMD; \citealt{geistTheo2019, vieillardLeve2021, xiaoConv, vaswaniGene2023, lanPoli2023, zhan2023policy}) has emerged as a unifying framework for common RL algorithms. The PMD policy update (e.g.\ \citealt{lanPoli2023}, Algorithm~1) 
\begin{equation}
    \label{eq::PMD}
    \begin{aligned}
         \pi_{k+1}(\cdot|s) = \arg\min_{\pi\in\Pi}\Bigl\{\IE_{a\sim \pi(\cdot|s)}[-Q_\alpha^{\pi_k}(s,a)]+ \alpha\,h(\pi(\cdot|s)) + \eta_k\, B_\omega(\pi,\pi_k;s)\Bigr\},
    \end{aligned}
\end{equation}
where $\eta_k > 0$ is a sequence of step sizes, $h$ some convex function and $B_\omega$ some Bregman Divergence, which acts as a trust-region term that prevents large deviations from the previous policy. SAC is thus an instance of PMD, with $h$ the negative Shannon entropy and no trust-region term, and the design choices studied here should transfer to other PMD algorithms, including those built on more general convex regularizers~\citep{leeTsal2019, alfanolearning}. They are also complementary to efforts that add trust-region terms to SAC~\citep{tomarMirr2021, neumann2025investigating}.

\textbf{Multi-step returns \& advantages in maximum-entropy RL.} Soft advantage estimators are not new: \citet{huang2020soft} introduced a variant of GAE for entropy-regularized objectives, and \citet{choeMaxi} use a closely related soft GAE in PPO-style algorithms, whereas we integrate soft advantages into SAC. Multi-step targets have likewise been studied in entropy-regularized settings~\citep{shilImpr2023}, including for transformer-based policies~\citep{tian2025chunking}; \citet{huangOffP2021} and \citet{lyskawaSACn2025} train SAC on replayed $\lambda$- and $n$-step returns, corrected by importance sampling and truncated importance sampling, respectively. \citet{kozuno2021revisiting} show that Peng's Q($\lambda$), which applies no correction, often outperforms conservative operators in continuous control, and characterize when its bias is benign; for SAC, however, they drop the entropy bonus from all but the last step of the target for stability. Our results add to these findings: the full soft $\lambda$-return, entropy terms included and with a learned temperature, trains stably from replay across a distribution of hyperparameters.

More broadly, recent work revisits core design assumptions in SAC, particularly regarding critic learning and target construction~\citep{shilImpr2023, zhouRevi2024, asadRevi2025}; we complement it with a systematic study of soft advantages, multi-step targets and replay within a unified SAC framework.

\section{Conclusion}
\label{sec:Conclusion}

We asked which choices in SAC's critic influence its outcomes by keeping the policy objective fixed and varying three of these choices: the learned quantity, the rollout depth of the target, and experience replay. In total, we evaluated across six critic variants over an entire random hyperparameter search on fourteen tasks. Learning soft advantages instead of soft Q-values neither makes SAC more robust to its hyperparameters nor improves its performance: the advantage and dueling critics stay close to \SACb{} in the share of well-performing configurations and in transfer, and on continuous tasks \SACA{} even reaches a lower peak. Since SAC's policy gradient is unaffected by state-dependent shifts of the critic, this is a finding about critic learning: in our setting, fitting the state-dependent level of the value is not what limits SAC's critic. The soft GAE critic, in turn, adds nothing over the simpler $\lambda$-return critic.

The clearest gain comes from training a $\lambda$-return critic on replayed segments. In continuous tasks, \SACLR{} holds the largest share of well-performing configurations, reaches the highest peak score at a lower relative sensitivity than \SACb{}, and keeps most of that peak under transfer across tasks. Its lead over the on-policy \SACL{} persists at a matched horizon, while neither a larger $\lambda$ nor a longer horizon helps within the sampled ranges; on discrete tasks, no variant improves on \SACb{}. In line with earlier evidence for uncorrected multi-step targets, the missing off-policy correction does not prevent these gains, and our results extend this observation to the full soft $\lambda$-return. In practice, SAC's critic parametrization can thus be chosen on grounds other than return, and multi-step targets can be paired with replay without correction.

Several limitations bound these conclusions. Our continuous-control evidence rests on a limited number of continuous action tasks, and given SAC's prevalence in that domain, further benchmarks are warranted. The replay comparison bounds the combined effect of target staleness, larger batches and a different update schedule rather than isolating the bias, and since \SACLR{} also differs from \SACb{} in its training regime, we attribute its gain to the combination rather than to the multi-step target alone. Any statement about a distribution over hyperparameters is conditional on the search space, which we centered on a configuration tuned for \SACb{}. Finally, we studied the variants in isolation from other recent SAC extensions to attribute effects cleanly; combining them remains open.

\subsection*{Acknowledgments}
This work was financially supported by Shell Information Technology International Limited and the Netherlands Enterprise Agency under the grant PPS23-3-03529461.

\bibliography{main}
\bibliographystyle{iclr2027_conference}

\newpage
\appendix

\section{Proofs}
\label{sec:appendix-proofs}

\subsection{Soft Advantages}

As highlighted in the related work, we can see the SAC policy update as an instance of Policy Mirror Descent (PMD; e.g. \citealt{zhan2023policy}). In fact, for all instances of PMD, the (soft) action-value function $Q_\alpha^\pi$ can be replaced by the corresponding (soft) advantage function.
\begin{equation}
    \begin{aligned}
        \pi_{k+1}(\cdot|s) &= \arg\min_{\pi\in\Pi}\left\{\IE_{a\sim\pi(\cdot|s)}[-Q^{\pi_k}_\alpha(s,a)]  + \alpha\,h(\pi(\cdot|s)) + \eta_k\, B_\omega(\pi,\pi_k;s)\right\} \\
        &= \arg\min_{\pi\in\Pi}\left\{\IE_{a\sim\pi(\cdot|s)}[-A^{\pi_k}_\alpha(s,a)] - \IE_{a\sim\pi_k(\cdot|s)}[Q^{\pi_k}_\alpha(s,a)]  + \alpha\,h(\pi(\cdot|s)) + \eta_k\, B_\omega(\pi,\pi_k;s)\right\}.
    \end{aligned}
\end{equation}
The term $\IE_{a\sim\pi_k(\cdot|s)}[Q^{\pi_k}_\alpha(s,a)]$ in the final expression can be dropped because it does not depend on $\pi$. This holds for any convex $h$ and is not limited to the negative Shannon entropy.

\textbf{Proof of Lemma~\ref{lem:2.1}.} We prove the two properties of Lemma~\ref{lem:2.1} together with a third, which relates soft and unregularized advantages.

\begin{lemma}[Lemma~\ref{lem:2.1}, extended]
    The soft advantages $A^\pi_\alpha$ satisfy the following properties for any $(s,a)\in\S\times\A$, $\alpha \geq 0$:
    \begin{enumerate}
        \item[(i)] $A^\pi_\alpha(s,a) = r(s,a) - \alpha h(\pi(\cdot|s)) + \gamma \IE_{s'\sim p(\cdot|s,a)}[V_\alpha^\pi(s')] - V^\pi_\alpha(s)$,
        \item[(ii)] $\IE_{a\sim\pi(\cdot|s)} [A^\pi_\alpha(s,a)] = 0$,
        \item[(iii)] $A^\pi_\alpha(s,a) = A^\pi(s,a) + \alpha H(\pi|s) - \alpha H(\pi|s,a)$, where $H(\pi|s,a)$ and $H(\pi|s)$ denote the cumulative discounted negative entropy starting with state-action pair $s,a$ and in state $s$, respectively:
        \begin{equation*}
            \begin{aligned}
                H(\pi|s,a) &:= \IE_{\tau\sim p_\pi|s_0=s,a_0=a}[\sum_{t\geq 0}\gamma^t h(\pi(\cdot|s_t))] \\
                H(\pi|s) &:= \IE_{a\sim\pi(\cdot|s))}[H(\pi|s,a)]
            \end{aligned}
        \end{equation*}
    \end{enumerate}
\end{lemma}

\begin{proof}

    Recall that the soft Q-function is defined as 
    \begin{align}
        \label{eq::app-01}
        Q_\alpha^\pi(s,a) = r(s,a) + \gamma \IE_{s'\sim p(\cdot|s,a)}[V_\alpha^\pi(s')].
    \end{align}
    Additionally, we can re-express the relation between $Q_\alpha^\pi$ and $V_\alpha^\pi$ as
    \begin{equation}
        \label{eq::app-02}
        \IE_{a\sim \pi(\cdot|s)}[Q_\alpha^\pi(s,a)] = V_\alpha^\pi(s) + \alpha h(\pi(\cdot|s)).
    \end{equation}

    (i) Now, we can insert the relations of Eqs. \ref{eq::app-01} and \ref{eq::app-02} into the definition of the soft advantages
    \begin{equation}
        \begin{aligned}
            A_\alpha^\pi(s,a) &= Q_\alpha^\pi(s,a) - \IE_{a\sim \pi(\cdot|s)}[Q_\alpha^\pi(s,a)] \\
            &= r(s,a) + \gamma \IE_{s'\sim p(\cdot|s,a)}[V_\alpha^\pi(s')] - V_\alpha^\pi(s) - \alpha h(\pi(\cdot|s)).
        \end{aligned}
    \end{equation}

    (ii) follows from definition as
    \begin{equation}
        E_{a\sim \pi(\cdot|s)}[A^\pi_\alpha(s,a)] = E_{a\sim \pi(\cdot|s)}[Q^\pi_\alpha(s,a)] - E_{a\sim \pi(\cdot|s)}[Q^\pi_\alpha(s,a)] = 0.
    \end{equation}

    (iii) By definition again, we can express 
    \begin{equation}
        V_\alpha^\pi(s) = V^\pi(s) - \alpha H(\pi|s).
    \end{equation}
    Inserting this into the expression obtained in (i) yields (after some reorganization of terms)
    \begin{align}
        A_\alpha^\pi(s,a) &= r(s,a) + \gamma \IE_{s'\sim p(\cdot|s,a)}[V_\alpha^\pi(s')] - V_\alpha^\pi(s) - \alpha h(\pi(\cdot|s))\\
        &= r(s,a) + \gamma \IE_{s' \sim p(\cdot|s,a)}[V^\pi(s')] - V^\pi(s) \\
        &\hspace{0.4cm}+ \alpha H(\pi|s) \\
        &\hspace{0.4cm}- \alpha\,\gamma\, \IE_{s' \sim p(\cdot|s,a), a'\sim \pi(\cdot|s')}[H(\pi|s',a')] - \alpha h(\pi(\cdot|s)) \\
        &= A^\pi(s,a) + \alpha H(\pi|s) - \alpha H(\pi|s,a).
    \end{align}
    
\end{proof}

Notably, we can also immediately see that if we use the soft value function as baseline subtraction, i.e. $\Tilde{A}^\pi_\alpha(s,a) := Q_\alpha^\pi(s,a) - V_\alpha^\pi(s)$, we obtain
\begin{align}
    \IE_{a\sim\pi(\cdot|s)}[\Tilde{A}^\pi_\alpha(s,a)] &= E_{a\sim \pi(\cdot|s)}[Q^\pi_\alpha(s,a)] - V_\alpha^\pi(s) = \alpha h(\pi(\cdot|s)) \tag*{(by Eq. \ref{eq::app-02})},
\end{align}
which does not vanish for non-deterministic policies in the regularized case ($\alpha > 0$).

\subsection{Soft n-step returns}

Though soft $n$-step returns have been used before with SAC \citep{shilImpr2023, lyskawaSACn2025, tian2025chunking}, we give the details behind Eqs.~\ref{eq::n-step Returns} and~\ref{eq::n-step Returns recursive} for completeness. Throughout, we write $\IE_\pi Q(s) := \IE_{a\sim\pi(\cdot|s)}[Q(s,a)]$ and $\mathcal{H}(s) := \mathcal{H}(\pi(\cdot|s))$.

\textbf{Claim} The n-step soft Bellman Operator satisfies the following expression:

\begin{equation*}
    (T_\alpha^\pi)^n Q(s_t,a_t) =\mathbb{E}_{\tau|s_t,a_t}\!\left[\sum_{k=0}^{n-1}\gamma^k r_{t+k} - \alpha \sum_{k=1}^{n}\gamma^k \log \pi(a_{t+k}\mid s_{t+k}) + \gamma^n Q(s_{t+n},a_{t+n})\right].
\end{equation*}

\begin{proof}
    Let $\pi$ be a fixed policy and define the soft Bellman operator
    \begin{equation*}
        (T_\alpha^\pi Q)(s,a) = r(s,a) + \gamma \,(P^\pi (Q - \alpha \log \pi))(s,a),
    \end{equation*}
    where $(P^\pi f)(s,a) := \mathbb{E}_{s' \sim p(\cdot\mid s,a),\, a' \sim \pi(\cdot\mid s')}\big[f(s',a')\big]$ and $\log \pi(s,a) := \log \pi(a\mid s)$.

    \vspace{0.2cm}
    We proceed by induction on $n$.
    
    \emph{Base case ($n=1$).}
    By definition of the soft Bellman operator,
    \begin{equation*}
        (T_\alpha^\pi Q)(s_t,a_t) = r_t+\gamma \mathbb{E}_{\tau|s_t,a_t}\!\left[Q(s_{t+1},a_{t+1})-\alpha \log \pi(a_{t+1}\mid s_{t+1})\right],
    \end{equation*}
    which coincides with the claimed expression for $n=1$.

    \emph{Inductive step.}
        Assume the statement holds for some $n \ge 1$, i.e. 
        \begin{equation*}
            (T_\alpha^\pi)^n Q(s_t,a_t) = \mathbb{E}_{\tau|s_t,a_t}\!\left[ \sum_{k=0}^{n-1}\gamma^k r_{t+k} - \alpha \sum_{k=1}^{n}\gamma^k \log \pi(a_{t+k}\mid s_{t+k}) + \gamma^n Q(s_{t+n},a_{t+n}) \right].
        \end{equation*}
        Consider $n+1$. Using the operator definition,
        \begin{equation*}
            (T_\alpha^\pi)^{n+1} Q(s_t,a_t) = r_t + \gamma \mathbb{E}_{\tau|s_t,a_t}\!\left[ (T_\alpha^\pi)^n Q(s_{t+1},a_{t+1}) - \alpha \log \pi(a_{t+1}\mid s_{t+1}) \right].
        \end{equation*}

        Applying the induction hypothesis at time $t+1$ and using the law of total expectation yields
        \begin{equation*}
            \begin{aligned}
                (T_\alpha^\pi)^{n+1} Q(s_t,a_t) &= \mathbb{E}_{\tau|s_t,a_t}\!\Bigg[ r_t + \sum_{k=0}^{n-1}\gamma^{k+1} r_{t+1+k} \\
                &\hspace{1cm}- \alpha\!\left( \gamma \log \pi(a_{t+1}\mid s_{t+1}) + \sum_{k=1}^{n}\gamma^{k+1}\log \pi(a_{t+1+k}\mid s_{t+1+k}) \right) \\
                &\hspace{3.8cm} + \gamma^{n+1} Q(s_{t+n+1},a_{t+n+1})\Bigg].
            \end{aligned}
        \end{equation*}
        
        Re-indexing the sums gives
        \begin{equation*}
            (T_\alpha^\pi)^{n+1} Q(s_t,a_t) = \mathbb{E}_{\tau|s_t,a_t}\!\left[\sum_{k=0}^{n}\gamma^k r_{t+k} - \alpha \sum_{k=1}^{n+1}\gamma^k \log \pi(a_{t+k}\mid s_{t+k}) + \gamma^{n+1} Q(s_{t+n+1},a_{t+n+1}) \right],
        \end{equation*}
        which is exactly the desired form. This completes the induction.
\end{proof}

Along a trajectory sampled from $\pi$, $a_{t+k}\sim\pi(\cdot|s_{t+k})$, so by the tower property each term $-\alpha\log\pi(a_{t+k}|s_{t+k})$ can be replaced by its conditional expectation $\alpha\mathcal{H}(s_{t+k})$, and the bootstrap $Q(s_{t+n},a_{t+n})$ by $\IE_\pi Q(s_{t+n})$, without changing the expectation:
\begin{equation*}
    (T_\alpha^\pi)^n Q(s_t,a_t) = \mathbb{E}_{\tau|s_t,a_t}\!\left[\sum_{k=0}^{n-1}\gamma^k r_{t+k} + \alpha \sum_{k=1}^{n}\gamma^k \mathcal{H}(s_{t+k}) + \gamma^n \IE_\pi Q(s_{t+n})\right].
\end{equation*}
This motivates the soft $n$-step return of Eq.~\ref{eq::n-step Returns} as an unbiased estimate of the $n$-step soft Bellman operator,
\begin{equation}
    \label{eq:helper1}
    R_{\alpha,t}^{(n)} = \sum_{k=0}^{n-1}\gamma^k \bigl\{r_{t+k} + \gamma\alpha\mathcal{H}(s_{t+k+1})\bigr\} + \gamma^n \IE_\pi Q(s_{t+n}).
\end{equation}
Unlike in \citet{tian2025chunking}, the entropy enters as a discounted bonus for the next state. This matches the soft Bellman operator, since $R_{\alpha,t}^{(1)} = r_t + \gamma\bigl(\IE_\pi Q(s_{t+1}) + \alpha\mathcal{H}(s_{t+1})\bigr)$ is the one-step target of Eq.~\ref{eq:2}. Based on Eq.~\ref{eq:lambdareturnoperator}, we further define the $\lambda$-return for a trajectory of length $N$ as
\begin{equation*}
    R_{\alpha,t}^\lambda := (1-\lambda) \sum_{n=1}^{N-1}\lambda^{n-1} R_{\alpha,t}^{(n)} + \lambda^{N-1} R_{\alpha,t}^{(N)}.
\end{equation*}

While the recursive relation in Eq.~\ref{eq::n-step Returns recursive} was used before, e.g.\ by \citet{huangOffP2021}, we provide a formal proof for completeness.

\textbf{Claim} The soft $\lambda$-returns satisfy the recursion
\begin{equation}
    \tag{\ref{eq::n-step Returns recursive}}
    R_{\alpha,t}^\lambda =
    \begin{cases}
        r_t & \text{if $s_{t+1}$ is terminal,}\\
        r_t + \gamma \bigl\{(1-\lambda)\IE_\pi Q(s_{t+1}) + \alpha\mathcal{H}(s_{t+1}) + \lambda R_{\alpha,t+1}^\lambda\bigr\} & \text{otherwise.}
    \end{cases}
\end{equation}

\begin{proof}
Write $r_t^{\alpha,\pi} := r_t + \gamma\alpha\mathcal{H}(s_{t+1})$. Then $R_{\alpha,t}^{(1)} = r_t^{\alpha,\pi} + \gamma\,\IE_\pi Q(s_{t+1})$ and, separating the first term, $R_{\alpha,t}^{(n)} = r_t^{\alpha,\pi} + \gamma R_{\alpha,t+1}^{(n-1)}$ for $n\geq 2$. Inserting both into the definition,
\begin{align*}
    R_{\alpha,t}^\lambda &= (1-\lambda)\bigl(r_t^{\alpha,\pi} + \gamma\,\IE_\pi Q(s_{t+1})\bigr) + (1-\lambda) \sum_{n=2}^{N-1} \lambda^{n-1} \bigl(r_t^{\alpha,\pi} + \gamma R_{\alpha,t+1}^{(n-1)}\bigr) + \lambda^{N-1} \bigl(r_t^{\alpha,\pi} + \gamma R_{\alpha,t+1}^{(N-1)}\bigr) \\
    &= r_t^{\alpha,\pi} + (1-\lambda)\gamma\,\IE_\pi Q(s_{t+1}) + \gamma\lambda \Bigl\{(1-\lambda) \sum_{n=1}^{N-2} \lambda^{n-1} R_{\alpha,t+1}^{(n)} + \lambda^{N-2} R_{\alpha,t+1}^{(N-1)} \Bigr\},
\end{align*}
where we grouped $r_t^{\alpha,\pi}$ and re-indexed the sum. The expression in braces is $R_{\alpha,t+1}^\lambda$, hence
\begin{equation*}
    R_{\alpha,t}^\lambda = r_t^{\alpha,\pi} + (1-\lambda)\gamma\,\IE_\pi Q(s_{t+1}) + \gamma\lambda R_{\alpha,t+1}^\lambda = r_t + \gamma \bigl\{(1-\lambda)\IE_\pi Q(s_{t+1}) + \alpha\mathcal{H}(s_{t+1}) + \lambda R_{\alpha,t+1}^\lambda\bigr\}.
\end{equation*}
By convention, $\IE_\pi Q(s_{t+1}) = 0$, $R_{\alpha,t+1}^\lambda = 0$ and $r_t^{\alpha,\pi} = r_t$ if $s_{t+1}$ is terminal, which yields the stated recursion.
\end{proof}

Algorithm~\ref{alg:PSAC} computes the same recursion in the equivalent form $R_{\alpha,t}^\lambda = r_t + \gamma\{(1-\lambda)\bar{V}(s_{t+1}) + \lambda(R_{\alpha,t+1}^\lambda + \alpha\mathcal{H}(s_{t+1}))\}$, with the soft value $\bar{V}(s) = \IE_\pi Q(s) + \alpha\mathcal{H}(s)$.

\subsection{GAE for soft advantages}

As in the unregularized case ($\alpha=0$), for $\lambda=1$ the soft GAE recovers a $(T-t)$-step lookahead minus the value of the current state:
\begin{align*}
    A^{\lambda=1}_\alpha(s_t,a_t) &= \sum_{i=0}^{T-t-1} \gamma^i \bigl\{r(s_{t+i},a_{t+i}) - \alpha h(\pi(\cdot|s_{t+i})) + \gamma V_\alpha(s_{t+i+1}) - V_\alpha(s_{t+i})\bigr\} \\
    &= \sum_{i=0}^{T-t-1} \gamma^i \bigl\{r(s_{t+i},a_{t+i}) - \alpha h(\pi(\cdot|s_{t+i}))\bigr\} + \gamma^{T-t} V_\alpha(s_T) - V_\alpha(s_t) \tag*{(telescoping)}\\
    &= R_{\alpha,t}^{(T-t)} - V_\alpha(s_t) - \alpha h(\pi(\cdot|s_t)),
\end{align*}
where the last step holds for $h=-\mathcal{H}$, with $Q = Q^\pi_\alpha$ in Eq.~\ref{eq:helper1} and $V_\alpha(s_T) = \IE_\pi Q(s_T) + \alpha\mathcal{H}(s_T)$.

\newpage
\section{Additional Results}
\label{appendix:additional-plots}

This appendix complements the main results with a variance decomposition, per-task rankings, probabilities of improvement at matched horizon, random-search budget curves, the $\lambda$ ablation and per-task performance profiles.

\subsection{Variance Decomposition}
\label{app:sec:VD}

An interesting question is whether the design choices change where performance variance originates. Performance varies with the task, the configuration, the training seed and the evaluation episode, and we decompose the total sum of squares accordingly. With $S_m$ training seeds on task $m$, $n_m = H S_m S'$, $x_{h,m,s} = \frac{1}{S'}\sum_{s'} x_{h,m,s,s'}$, $x_{h,m}$ as in Section~\ref{sec:Experiments}, $\mu_m = \frac{1}{H}\sum_h x_{h,m}$ and $\mu = \frac{1}{M}\sum_m \mu_m$,
\begin{equation}
    \begin{aligned}
        SS_{\text{task}}    &= \sum_m n_m(\mu_m-\mu)^2, \\
        SS_{\text{config}}  &= \sum_m S_m S' \sum_h \bigl(x_{h,m}-\mu_m\bigr)^2,\\
        SS_{\text{seed}}    &= \sum_m S' \sum_{h,s}\bigl(x_{h,m,s}-x_{h,m}\bigr)^2, \\
        SS_{\text{episode}} &= \sum_{h,m,s,s'}\bigl(x_{h,m,s,s'}-x_{h,m,s}\bigr)^2,
    \end{aligned}
\end{equation}
and each share is $SS_\ell \big/ \sum_{\ell'} SS_{\ell'}$.

\begin{figure}[h]
  \centering
  \includegraphics[width=0.95\textwidth]{images_v2/legend_discrete.pdf}
  \\[0.3em]
  \begin{subfigure}{0.24\textwidth}\includegraphics[width=\linewidth]{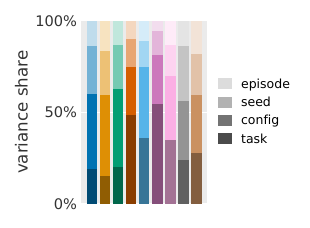}\caption{Discrete Tasks \\ all configurations}\end{subfigure}\hfill
  \begin{subfigure}{0.24\textwidth}\includegraphics[width=\linewidth]{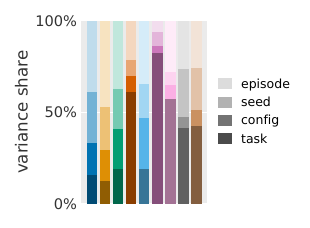}\caption{Discrete Tasks \\ top-quartile configurations}\end{subfigure}\hfill
  \begin{subfigure}{0.24\textwidth}\includegraphics[width=\linewidth]{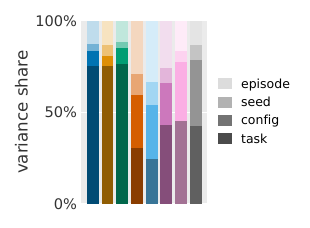}\caption{Continuous Tasks \\ all configurations}\end{subfigure}\hfill
  \begin{subfigure}{0.24\textwidth}\includegraphics[width=\linewidth]{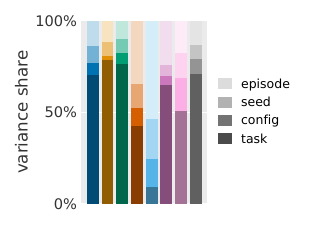}\caption{Continuous Tasks \\ top-quartile configurations}\end{subfigure}
  \caption{\textbf{Variance decomposition.} Share of total variance attributable to task, configuration, training seed and evaluation episode, over all configurations (a, c) and over each method's top-quartile configurations (b, d), on discrete (a, b) and continuous (c, d) tasks.}
    \label{fig:VD01}
\end{figure}

The resulting profiles are shown in Figures~\ref{fig:VD01}. On continuous tasks the one-step methods draw about $75\%$ of their variance from differences between tasks, whereas the multi-step variants draw only $24$--$45\%$, with correspondingly more coming from configuration, seed and episode noise. A smaller between-task share means performance depends less on which task happens to be drawn, which is consistent with the stronger transfer reported above. On discrete tasks the pattern reverses, and what distinguishes the multi-step variants there is instead a smaller training-noise share at comparable evaluation noise.

The comparison is biased by performance level, since a method that fails uniformly has little seed-to-seed variance left to decompose. Restricting the decomposition to each method's top-quartile configurations (Fig.~\ref{fig:VD01}b,d) shrinks the configuration share and raises the seed and episode shares, as expected, but leaves the between-task trend mostly intact.

\subsection{Per Task Ranking}
\label{sec:app:pertaskranking}

\begin{figure}[t]
    \centering
    \includegraphics[width=0.75\linewidth]{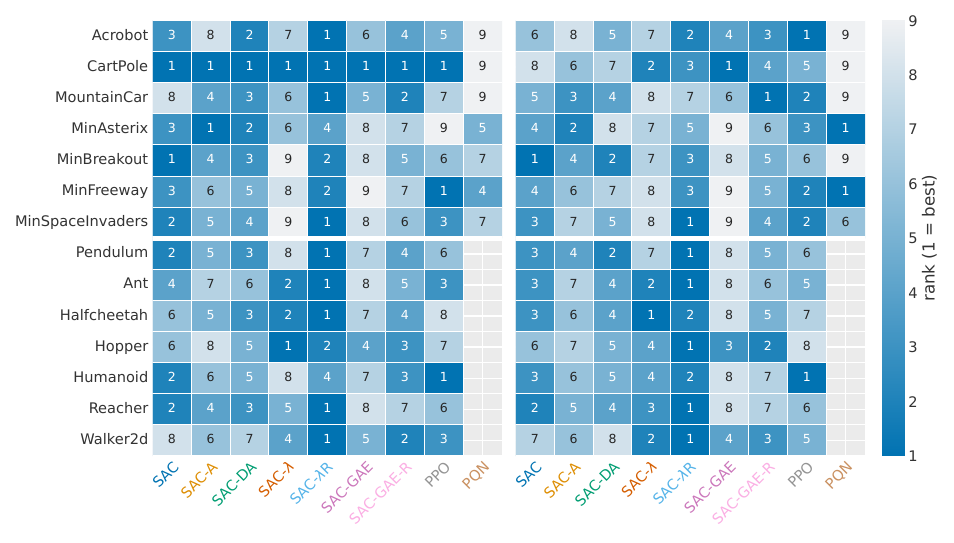}
    \caption{\textbf{Per-task ranks} (1 = best). Left: by peak score $\max_h x_{h,m}$; right: by $\mathrm{IQM}_h(x_{h,m})$. PQN is not defined for continuous action spaces.}
    \label{fig:App03}
\end{figure}

Due to the amount of environment-algorithm combinations, most of our analysis has been focused on aggregate metrics. To complement this view Figure~\ref{fig:App03} shows the per-task ranks of all methods, by peak score $\max_h x_{h,m}$ (left) and by the interquartile mean over configurations, $\mathrm{IQM}_h(x_{h,m})$ (right), a proxy for performance under limited tuning. By peak score, \SACLR{} ranks first on nine of the fourteen tasks, including five of the seven continuous ones, and never below fourth. By IQM the picture is similar but less pronounced: \SACLR{} ranks first or second on every continuous task, but drops to seventh on MountainCar.

\subsection{Random-Search Budget Curve}

\begin{figure}
    \centering
  \includegraphics[width=0.95\textwidth]{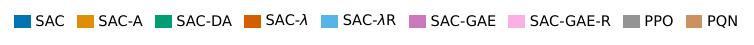}
  \\[0.3em]
  \begin{subfigure}{0.35\textwidth}\includegraphics[width=\linewidth]{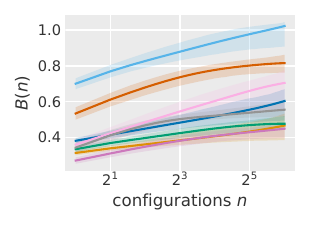}\caption{Continuous}\end{subfigure}
  \begin{subfigure}{0.35\textwidth}\includegraphics[width=\linewidth]{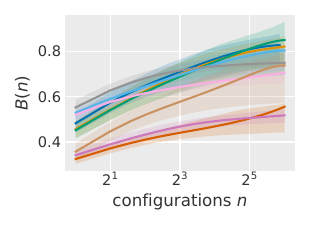}\caption{Discrete}\end{subfigure}
  \caption{\textbf{Random-search budget curves.} Expected normalized score of the best shared configuration among $n$ sampled ones. Bands are 95\% bootstrap confidence intervals.}
    \label{fig:App04}
\end{figure}

Ranking by the IQM in Section \ref{sec:app:pertaskranking} was used as a proxy for how rankings change with limited tuning. To further investigate that effect, we use budget curves~\citep{dodge2019show}: they give the expected score of the best shared configuration among $n$ randomly drawn ones,
\begin{equation}
    B(n) = \IE_{\mathcal{S}\subseteq\{1,\dots,H\},\,|\mathcal{S}|=n}\Bigl[\max_{h \in \mathcal{S}}\frac{1}{M} \sum_{m=1}^M x_{h,m}\Bigr].
\end{equation}
On continuous tasks (Figure~\ref{fig:App04}a), \SACLR{} leads at every budget and \SACL{} is second, so their advantage does not depend on how much tuning is spent. On discrete tasks (Figure~\ref{fig:App04}b), \SACb{}, the advantage variants and the replay-based variants overlap almost fully and gain similarly from tuning, the on-policy multi-step variants stay below, and PQN gains most from additional configurations.

\newpage
\subsection{Replay at Matched Horizon: Probability of Improvement}
\label{app:sec:poi}

\begin{figure}[t]
    \centering
    \begin{subfigure}{0.24\textwidth}\includegraphics[width=\linewidth]{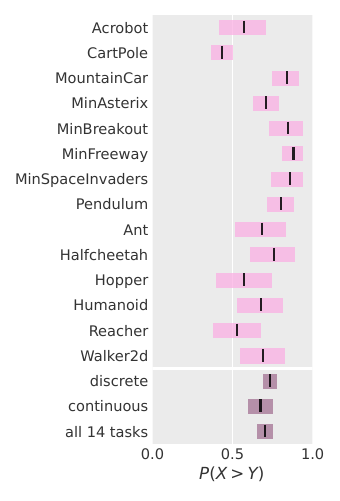}\caption{X=\SACGR{}, \\ Y=\SACG{}}\end{subfigure}\hspace{0.24\textwidth}
  \begin{subfigure}{0.24\textwidth}\includegraphics[width=\linewidth]{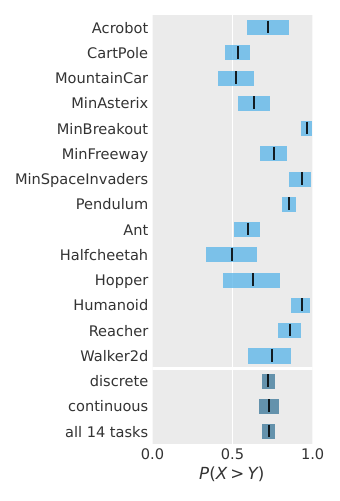}\caption{X=\SACLR{},\\  Y=\SACL{}}\end{subfigure}
    \caption{\textbf{Probability of improvement at $N=32$} of the replay-based over the on-policy variant, per task and aggregated. (a): \SACGR{} vs.\ \SACG{}; (b): \SACLR{} vs.\ \SACL{}; the dashed line marks $0.5$. Bars are 95\% stratified bootstrap confidence intervals.}
    \label{fig:App02}
\end{figure}

To complement Figure~\ref{fig-exp:2}c,d, we compare each replay-based variant with its on-policy counterpart at the shared horizon $N=32$, using the probability of improvement of \citet{agarwalDeep2021}, extended to configurations and evaluation seeds. For methods $X$ and $Y$,
\begin{equation}
    \begin{aligned}
        P(X > Y) &= \frac{1}{M} \sum_{m=1}^M P_m(X > Y), \\
        P_m(X > Y) &= \frac{1}{(H S S')^2} \sum_{h,s,s'}\,\sum_{\tilde{h},\tilde{s},\tilde{s}'} \Bigl(\mathbbm{1}_{\{x_{h,m,s,s'} > y_{\tilde{h},m,\tilde{s},\tilde{s}'}\}} + \tfrac{1}{2}\mathbbm{1}_{\{x_{h,m,s,s'} = y_{\tilde{h},m,\tilde{s},\tilde{s}'}\}}\Bigr),
    \end{aligned}
\end{equation}
where the sums run over the configurations sampled with $N=32$. For both replay-base variants the CI for the probability of improvement does not contain zero on either suite (Figure~\ref{fig:App02}a,b), demonstrating a statistically significant improvement.

\subsection{\texorpdfstring{$\lambda$}{lambda} Ablations}
\label{app:sec:lambdaablation}

\begin{figure}[t]
    \centering
    \includegraphics[width=\linewidth]{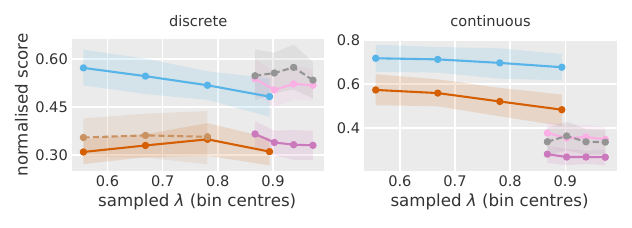}
    \includegraphics[width=\linewidth]{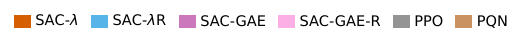}
    \caption{\textbf{Effect of the sampled $\lambda$.} normalized score against $\lambda$, binned at bin centers, for every method that samples it. The $\lambda$-return critics are sampled on $[0.50,0.95]$ and the GAE critics on $[0.85,0.99]$, so the two families occupy different parts of the axis and should not be compared by slope across families. Bands are 95\% bootstrap confidence intervals; slopes are given in Table~\ref{tab:lambda-slopes}.}
    \label{fig:App06}
\end{figure}

$\lambda$ is sampled for every multi-step variant, independently of all other hyperparameters, so the score against $\lambda$ within one variant is a genuine ablation of the multi-step target: same architecture, data and update schedule, but a longer effective horizon (Figure~\ref{fig:App06}). We quantify it by two weighted least-squares slopes of the task-averaged score of a configuration, $\frac{1}{M}\sum_m x_{h,m}$, in $\lambda$ (Table~\ref{tab:lambda-slopes}). The marginal slope regresses on $\lambda$ alone and thus averages over all other hyperparameters; the partial slope is the $\lambda$ coefficient in a regression on all hyperparameters sampled for the method, with learning rates, sizes and counts on a log scale. Since $\lambda$ is sampled independently of the other hyperparameters, both estimate the same effect, but the partial slope is more precise, as the controls absorb variance that $\lambda$ does not explain. Of the sixteen estimates, only two exclude zero, and both are negative: within the sampled ranges, placing more weight on long-range returns does not improve performance.

\begin{table}[h]
\centering
\caption{Effect of $\lambda$ on performance. Slopes of the task-averaged normalized score of a configuration (in percentage points) per $+0.1$ in $\lambda$, from weighted least squares. The marginal slope regresses on $\lambda$ alone; the partial slope is the $\lambda$ coefficient in a regression on all hyperparameters sampled for the method, with learning rates, sizes and counts on a log scale. For the GAE critics, $\lambda$ is the GAE coefficient, sampled on $[0.85, 0.99]$; the $\lambda$-return critics sample $\lambda \in [0.50, 0.95]$. Lower and upper bounds are 95\% bootstrap confidence intervals; bold marks intervals that exclude zero.}
\label{tab:lambda-slopes}
\small
\setlength{\tabcolsep}{5pt}
\begin{tabular}{l r r r r r r}
\toprule
 & \multicolumn{3}{c}{Marginal} & \multicolumn{3}{c}{Partial} \\
\cmidrule(lr){2-4}\cmidrule(lr){5-7}
Method & slope & lower & upper & slope & lower & upper \\
\midrule
\multicolumn{7}{l}{\emph{Discrete}} \\
\SACL{}  & $+0.3$ & $-1.3$ & $+2.0$ & $+0.3$ & $-1.1$ & $+1.8$ \\
\SACLR{} & $-2.0$ & $-4.1$ & $+0.1$ & $\mathbf{-2.2}$ & $-3.6$ & $-0.6$ \\
\SACG{}  & $-2.6$ & $-8.2$ & $+3.1$ & $-3.2$ & $-6.9$ & $+0.4$ \\
\SACGR{} & $-0.8$ & $-7.3$ & $+5.2$ & $-0.9$ & $-5.5$ & $+3.5$ \\
\midrule
\multicolumn{7}{l}{\emph{Continuous}} \\
\SACL{}  & $\mathbf{-2.7}$ & $-5.3$ & $-0.1$ & $-1.4$ & $-3.2$ & $+0.5$ \\
\SACLR{} & $-1.1$ & $-3.3$ & $+1.1$ & $-1.4$ & $-3.1$ & $+0.2$ \\
\SACG{}  & $-1.4$ & $-6.0$ & $+3.1$ & $+0.5$ & $-3.5$ & $+4.1$ \\
\SACGR{} & $-3.5$ & $-10.0$ & $+3.3$ & $-1.1$ & $-6.5$ & $+4.8$ \\
\bottomrule
\end{tabular}
\end{table}

\newpage
\subsection{Per Environment Performance Profiles}

\begin{figure}[htbp]
  \centering
  \begin{subfigure}{0.24\textwidth}\includegraphics[width=\linewidth]{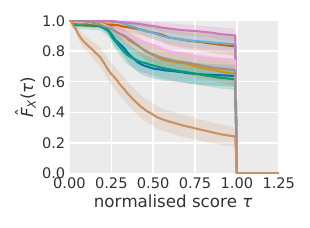}\caption{CartPole}\end{subfigure}\hfill
  \begin{subfigure}{0.24\textwidth}\includegraphics[width=\linewidth]{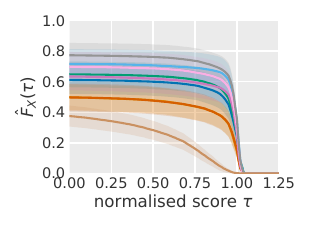}\caption{Acrobot}\end{subfigure}\hfill
  \begin{subfigure}{0.24\textwidth}\includegraphics[width=\linewidth]{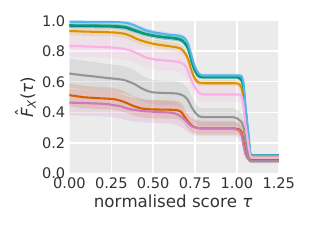}\caption{Pendulum}\end{subfigure}\hfill
  \begin{subfigure}{0.24\textwidth}\includegraphics[width=\linewidth]{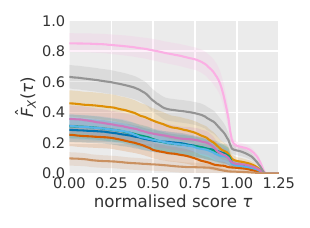}\caption{MountainCar}\end{subfigure}\hfill
  \\[1ex]
  \begin{subfigure}{0.24\textwidth}\includegraphics[width=\linewidth]{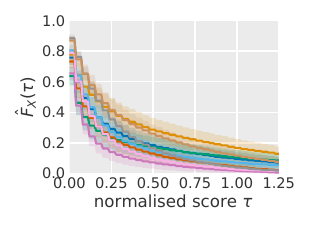}\caption{MinAsterix}\end{subfigure}\hfill
  \begin{subfigure}{0.24\textwidth}\includegraphics[width=\linewidth]{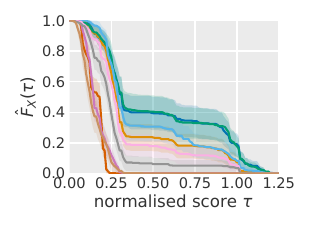}\caption{MinBreakout}\end{subfigure}\hfill
  \begin{subfigure}{0.24\textwidth}\includegraphics[width=\linewidth]{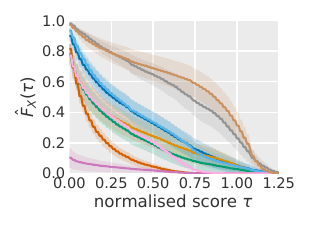}\caption{MinFreeway}\end{subfigure}\hfill
  \begin{subfigure}{0.24\textwidth}\includegraphics[width=\linewidth]{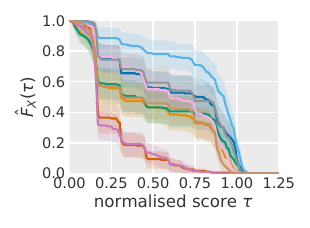}\caption{MinSpaceInvaders}\end{subfigure}\hfill
  \\[1ex]
  \begin{subfigure}{0.24\textwidth}\includegraphics[width=\linewidth]{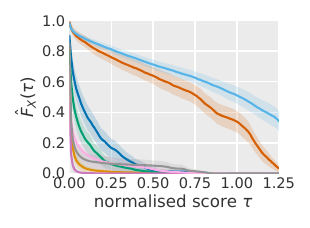}\caption{Ant}\end{subfigure}\hfill
  \begin{subfigure}{0.24\textwidth}\includegraphics[width=\linewidth]{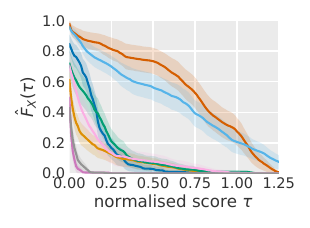}\caption{Halfcheetah}\end{subfigure}\hfill
  \begin{subfigure}{0.24\textwidth}\includegraphics[width=\linewidth]{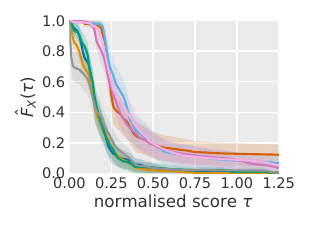}\caption{Hopper}\end{subfigure}\hfill
  \begin{subfigure}{0.24\textwidth}\includegraphics[width=\linewidth]{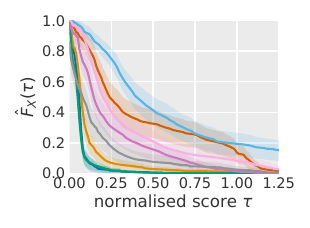}\caption{Walker2d}\end{subfigure}\hfill
  \\[1ex]
  \begin{subfigure}{0.24\textwidth}\includegraphics[width=\linewidth]{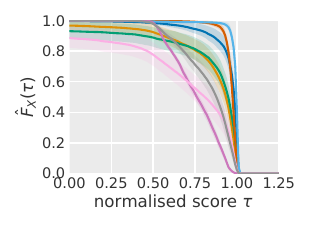}\caption{Reacher}\end{subfigure}
  \begin{subfigure}{0.24\textwidth}\includegraphics[width=\linewidth]{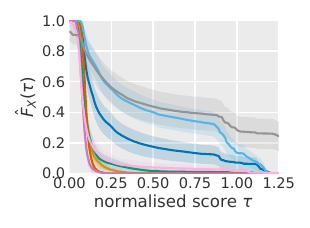}\caption{Humanoid}\end{subfigure}\hspace{0.24\textwidth}
  \\[1ex]
  \begin{subfigure}{\textwidth}\includegraphics[width=\linewidth]{images_v2/legend_discrete.pdf}\end{subfigure}
  \caption{\textbf{Per-task performance profiles}: share of configurations and seeds reaching normalized score $\tau$ on each task (Eq.~\ref{eq:score-distr} with $M=1$).}
\end{figure}

\newpage
\section{Experiment Details}
\label{appendix:experiment-details}

\subsection{Random Search Protocol}
\label{app:random-search}
For each of the $M$ environments, we perform a random hyperparameter search consisting of $H$ sampled configurations. For each configuration, we train an agent $S$ times with different random seeds. After training each agent, we additionally perform $S'=50$ evaluation rollouts per seed, yielding robust estimates of post-training performance. Hyperparameter configurations are drawn by Sobol sampling over the ranges specified in Table~\ref{tab:sweep-ranges}. All remaining hyperparameters are held fixed to the default SAC settings. The numbers for each sweep are specified in Table~\ref{tab:runs}.

Hyperparameters that exist only in a subset of the variants are sampled on the same footing as the shared ones: $\lambda \in [0.50, 0.95]$ for the $\lambda$-return critics, the GAE coefficient in $[0.85, 0.99]$ for the GAE critics, and the horizon $N$, i.e.\ the rollout length for the on-policy variants and the replayed segment length for the replay-based ones.

\begin{table}[h]
\centering
\caption{Configurations per task ($H$; minimum and maximum across the tasks of a suite), training seeds per configuration ($S$) and resulting training runs per suite.}
\small
\setlength{\tabcolsep}{4pt}
\begin{tabular}{l rrr rrr rrr rrr}
\toprule
 & \multicolumn{3}{c}{$H_{\min}$} & \multicolumn{3}{c}{$H_{\max}$} & \multicolumn{3}{c}{$S$} & \multicolumn{3}{c}{Training runs} \\
\cmidrule(lr){2-4} \cmidrule(lr){5-7} \cmidrule(lr){8-10} \cmidrule(lr){11-13}
Method & Brax & CC & MinAtar & Brax & CC & MinAtar & Brax & CC & MinAtar & Brax & CC & MinAtar \\
\midrule
PPO       & 64 & 64 & 64 & 64 & 64 & 64 & 3 & 5 & 3 & 1152 & 1280 & 768 \\
PQN       & -- & 64 & 64 & -- & 64 & 64 & --& 5 & 3 & --   &  960 & 768 \\
SAC       & 64 & 58 & 64 & 64 & 65 & 64 & 3 & 5 & 3 & 1152 & 1255 & 768 \\
SAC-A     & 64 & 64 & 64 & 64 & 64 & 64 & 3 & 5 & 3 & 1152 & 1280 & 768 \\
SAC-DA    & 64 & 64 & 64 & 64 & 64 & 64 & 3 & 5 & 3 & 1152 & 1280 & 768 \\
SAC-GAE   & 64 & 64 & 64 & 64 & 64 & 64 & 3 & 5 & 3 & 1152 & 1280 & 768 \\
SAC-GAE-R & 64 & 64 & 64 & 64 & 64 & 64 & 3 & 5 & 3 & 1152 & 1280 & 768 \\
\SACL{}     & 64 & 64 & 64 & 64 & 64 & 64 & 3 & 5 & 3 & 1152 & 1280 & 768 \\
\SACLR{}   & 64 & 64 & 64 & 64 & 64 & 64 & 3 & 5 & 3 & 1152 & 1280 & 768 \\
\bottomrule
\end{tabular}
\label{tab:runs}
\end{table}

Notably, the replay variants of \SACL{} and \SACG{} allow for a higher rollout length in our implementation. For \SACL{} and \SACG{} $N$ had to be short, so that there were a reasonable amount of actor updates.

\begin{table}[h]
\centering
\small
\caption{Sampled hyperparameter ranges. Continuous ranges are given as intervals, discrete ones
as sets; ``log'' denotes log-uniform sampling. }
\label{tab:sweep-ranges}
\begin{tabular}{ll}
\toprule
\textbf{Hyperparameter} & \\
\midrule
\multicolumn{2}{l}{\textit{Shared by all methods}} \\
Discount factor $\gamma$      & $[0.95,\,0.995]$  \\
Actor learning rate           & $[5\!\times\!10^{-5},\,3\!\times\!10^{-3}]$, log \\
Critic learning rate          & $[5\!\times\!10^{-5},\,3\!\times\!10^{-3}]$, log \\
Temperature learning rate     & $[5\!\times\!10^{-5},\,3\!\times\!10^{-3}]$, log \\
\midrule
\multicolumn{2}{l}{\textit{One-step replay critics (\SACb{}, \SACA{}, \SACDA{})}} \\
Batch size                    & $\{64,128,512,1024,2048\}$ \\
Replay buffer size            & $\{10^5,\,2.5\!\times\!10^5,\,10^6\}$ \\
Critic updates per iteration  & $\{4,8,16,32,64\}$ \\
\midrule
\multicolumn{2}{l}{\textit{On-policy multi-step critics (\SACL{}, \SACG{})}} \\
Rollout length $N$            & $\{8,\,16,\,32\}$  \\
Critic epochs per iteration   & $\{4,8,16,32\}$ \\
Critic minibatches            & $\{2,4,8,16\}$ \\
\midrule
\multicolumn{2}{l}{\textit{Replay-based multi-step critics (\SACLR{}, \SACGR{})}} \\
Replayed segment length $N$   & $\{32,\,64,\,128,\,256\}$  \\
Batch size                    & $\{1024,\,2048,\,4096\}$  \\
Replay buffer size & $\{10^5,\,2.5\!\times\!10^5,\,1.5\!\times\!10^6\}$ \\
Critic updates per iteration  & $\{4,8,16,32\}$ \\
\midrule
\multicolumn{2}{l}{\textit{Method-specific}} \\
$\lambda$ ($\lambda$-return critics)     & $[0.50,\,0.95]$  \\
$\lambda$ (GAE critics)                  & $[0.85,\,0.99]$  \\
Advantage-critic learning rate           & $[5\!\times\!10^{-5},\,3\!\times\!10^{-3}]$, log \\
Advantage-critic updates (SAC-A only)    & $\{4,8,16,32,64\}$\\
\bottomrule
\end{tabular}
\end{table}

\newpage
\subsection{Environment Return ranges}

Table~\ref{tab:refscores} lists the reference returns used to normalize scores (Section~\ref{sec:Experiments}).

\begin{table}[h]
\centering
\caption{Reference returns used for normalization. The floor is the minimum achievable return and maps to $0$; the $95$th percentile of per-run mean returns, pooled over all algorithms, maps to $1$. The pooled maximum is given for reference. Lower and upper bounds are 95\% bootstrap confidence intervals.}
\label{tab:refscores}
\small
\setlength{\tabcolsep}{4.5pt}
\begin{tabular}{l r r r r r r r}
\toprule
 & & \multicolumn{3}{c}{95th percentile} & \multicolumn{3}{c}{Maximum} \\
\cmidrule(lr){3-5}\cmidrule(lr){6-8}
Task & Floor & value & lower & upper & value & lower & upper \\
\midrule
\multicolumn{8}{l}{\emph{Discrete}} \\
CartPole         & $0$    & $500$    & $500$    & $500$    & $500$    & $500$    & $500$ \\
Acrobot          & $-500$ & $-78.0$  & $-78.5$  & $-77.5$  & $-65.6$  & $-69.4$  & $-65.6$ \\
MountainCar      & $-200$ & $-99.3$  & $-100.2$ & $-98.9$  & $-95.2$  & $-95.9$  & $-95.2$ \\
MinAsterix       & $0$    & $25.3$   & $22.9$   & $28.1$   & $57.4$   & $45.3$   & $57.4$ \\
MinBreakout      & $0$    & $93.4$   & $92.1$   & $95.0$   & $112.0$  & $107.7$  & $112.0$ \\
MinFreeway       & $0$    & $51.4$   & $49.6$   & $52.8$   & $60.5$   & $59.1$   & $60.5$ \\
MinSpaceInvaders & $0$    & $166$    & $165$    & $167$    & $176$    & $172$    & $176$ \\
\midrule
\multicolumn{8}{l}{\emph{Continuous}} \\
Pendulum         & $-500$ & $-143.3$ & $-143.9$ & $-142.7$ & $-137.9$ & $-139.0$ & $-137.9$ \\
Reacher          & $-400$ & $-8.03$  & $-8.79$  & $-7.59$  & $-4.94$  & $-5.82$  & $-4.94$ \\
Ant              & $0$    & $5940$   & $5728$   & $6130$   & $7911$   & $7368$   & $7911$ \\
HalfCheetah      & $0$    & $16100$  & $14800$  & $16900$  & $21400$  & $19900$  & $21400$ \\
Hopper           & $0$    & $2710$   & $2075$   & $3248$   & $4286$   & $3825$   & $4286$ \\
Humanoid         & $0$    & $5760$   & $5338$   & $6488$   & $9412$   & $8773$   & $9412$ \\
Walker2d         & $0$    & $4465$   & $4010$   & $4793$   & $8798$   & $7590$   & $8798$ \\
\bottomrule
\end{tabular}
\end{table}

\newpage
\section{Algorithm Details}
\label{sec:appendix-algdetails}

Following the main text, $\theta$ denotes critic parameters, $\phi$ value-network parameters and $\psi$ policy parameters; target networks are indicated by $\bar{\theta}$ and $\bar{\phi}$. $N$ denotes the horizon (rollout or segment length), as in the main text, and $C$ the period of the hard target-network updates.

\textbf{Implementation differences at a glance.} All variants share the policy objective of Eq.~\ref{eq:4}, the learned temperature with its target-entropy schedule (Remark~\ref{rk:SAC-1}) and the optimizer. Beyond the critic target, they differ as follows: (i) \SACb{}, \SACL{} and \SACLR{} use two critics (Remark~\ref{rk:SAC-2}), whereas \SACA{} and \SACDA{} use, by design, a single advantage and value network; (ii) the GAE variants use an unsquashed Gaussian policy, the score-function gradient of Eq.~\ref{eq:helper2} and batch-normalized advantages, which rescales the advantages relative to $\alpha$ and thus the effective temperature of their policy update; (iii) the on-policy variants take a single full-batch policy step per iteration, all others $K_\pi$ minibatch steps; (iv) the GAE variants use no target network.

\subsection{\SACb{}}

\begin{algorithm}[h]
\caption{Vanilla Soft Actor-Critic (SAC)}
\label{alg:SAC}
\begin{algorithmic}
\STATE {\bfseries Initialize:} critics $Q_{\theta_1}$, $Q_{\theta_2}$, target critics $Q_{\bar{\theta}_j} \leftarrow Q_{\theta_j}$, policy network $\pi_\psi$
\STATE Initialize learned temperature $\alpha = \exp(\log \alpha)$ and target-entropy schedule $\bar{\mathcal{H}}_t$
\STATE Initialize replay buffer $D \leftarrow \emptyset$ and fill it with $W$ warm-up transitions collected by $\pi_\psi$

\FOR{$t=1,\dots,\mathrm{budget}$}
    \STATE environment step: sample action $a \sim \pi_\psi(\cdot \mid s)$, next state $s' \sim p(\cdot \mid s,a)$
    \STATE Store transition $(s,a,r(s,a),d,s')$ in $D$, with $d$ the termination flag

    \IF{$t \bmod U = 0$}
        \STATE \textit{(Policy Evaluation Step)}
        \FOR{$k=1,\dots,K_Q$}
            \STATE Sample minibatch $B = \{(s_i,a_i,r_i,d_i,s'_i)\}_{i=1}^b$ from $D$
            \STATE Compute soft target
            \begin{equation*}
                T_i = r_i + (1-d_i)\, \gamma \, \mathbb{E}_{a' \sim \pi_\psi(\cdot \mid s'_i)} \big[ \operatorname*{agg}_{j=1,2} Q_{\bar{\theta}_j}(s_i',a')  - \alpha \log(\pi_\psi(a' \mid s'_i))\big] \tag*{cf. Eq. \ref{eq:2}}
            \end{equation*}
            \STATE For each critic calculate critic loss
            \begin{equation*}
                L^\mathrm{SAC}_\mathrm{critic}(\theta_j) = \frac{1}{b} \sum_{i=1}^b \left\{Q_{\theta_j}(s_i,a_i) - T_i\right\}^2 \tag*{cf. Eq. \ref{eq::SACLoss}}
            \end{equation*}
            \STATE Update critics via gradient descent on $\nabla_{\theta_j}L^\mathrm{SAC}_\mathrm{critic}$
            \STATE \textit{(Target network update)} $\bar{\theta}_j \leftarrow \tau \theta_j + (1-\tau)\bar{\theta}_j$
        \ENDFOR

        \STATE
        \STATE \textit{(Policy Improvement Step)}
        \FOR{$k=1,\dots,K_\pi$}
            \STATE Sample minibatch of states $\{s_i\}_{i=1}^b$ from $D$ and compute
            \begin{equation*}
                L_\pi(\psi) = \frac{1}{b} \sum_{i=1}^b
            \mathbb{E}_{a \sim \pi_\psi(\cdot \mid s_i)}
            \big[ -\operatorname*{agg}_{j=1,2} Q_{\theta_j}(s_i,a)
            + \alpha \log(\pi_\psi(a\mid s_i))\big] \tag*{cf. Eq. \ref{eq:4}}
            \end{equation*}
            \STATE Update policy via gradient descent on $\nabla_\psi L_\pi$
        \ENDFOR

        \STATE
        \STATE \textit{(Temperature Step)}
        \FOR{$k=1,\dots,K_\alpha$}
            \STATE Sample minibatch of states $\{s_i\}_{i=1}^b$ from $D$ and compute
            \begin{equation*}
                L(\alpha) = -\frac{1}{b} \sum_{i=1}^b \alpha \,
            \mathbb{E}_{a \sim \pi_\psi(\cdot \mid s_i)}
            \big[ \log(\pi_\psi(a\mid s_i)) + \bar{\mathcal{H}}_t \big]
            \end{equation*}
            \STATE Update temperature via gradient descent on $\nabla_{\log \alpha} L$
        \ENDFOR
    \ENDIF
\ENDFOR
\end{algorithmic}
\end{algorithm}

\begin{remark}[double Q function]
\label{rk:SAC-2}
    To reduce the bias induced by overestimation of the Q-values, we implement an ensemble of Q-networks, here of size two. The operator $\operatorname*{agg}_{j=1,2}$ aggregates the two critics and depends on the action space: for continuous actions we take the minimum, $\min_{j=1,2} Q_{\bar{\theta}_j}$, whereas for discrete actions we take the mean, $\frac{1}{2}\sum_{j=1,2} Q_{\bar{\theta}_j}$ \citep{zhou2022revisiting}. The same aggregation is used in the critic target (with target critics) and in the policy loss (with online critics).
\end{remark}

\vspace{0.1cm}
\begin{remark}[Discrete vs.\ continuous action spaces]
    \label{rk:SAC-3}
    The algorithm handles either discrete or continuous action spaces, but not mixed ones. For discrete actions, $\pi_\psi(\cdot \mid s)$ is a categorical distribution, and the critics output one value per action, $Q_{\theta_j}(s,\cdot) \in \mathbb{R}^{|\mathcal{A}|}$, from which $Q_{\theta_j}(s_i,a_i)$ is selected. All expectations over $a \sim \pi_\psi$ (in the critic target, the policy loss and the temperature loss) are then computed exactly as $\sum_{a \in \mathcal{A}} \pi_\psi(a \mid s)\,[\,\cdot\,]$. For continuous actions, $\pi_\psi$ is a tanh-squashed Gaussian, the critics take $(s,a)$ as input, and every expectation is estimated with a single reparameterized sample $a \sim \pi_\psi(\cdot \mid s)$.
\end{remark}

\vspace{0.1cm}
\begin{remark}[$\alpha$-schedule]
    \label{rk:SAC-1}
    The temperature is learned, parameterized as $\log \alpha$ and initialized at $\alpha_0 = 0.2$, following \citep{haarnojaSoft2019}. The target entropy is linearly annealed via a scale $\sigma_t$ that goes from $1$ to $0.1$ over the course of training: for discrete actions $\bar{\mathcal{H}}_t = 0.89\, \sigma_t \log|\mathcal{A}|$, and for continuous actions $\bar{\mathcal{H}}_t = -(2-\sigma_t)\dim(\mathcal{A})$. In both cases the target entropy decreases, driving the policy towards less stochastic behavior.
\end{remark}

\begin{remark}[Target network updates]
    \label{rk:SAC-6}
    We use hard target updates, $\bar{\theta} \leftarrow \theta$ every $C$ critic gradient steps, i.e. $\tau(c) = \mathbbm{1}_{\{C \mod N = 0\}}$ (not environment steps), which performed better than Polyak averaging in our implementation. The same holds for all variants with target networks.
\end{remark}

\newpage
\subsection{\SACA{}}

\begin{algorithm}[H]
\caption{SAC-A}
\label{alg:ASAC}
\begin{algorithmic}
\STATE {\bfseries Initialize:} advantage critic $A_\theta$, soft value network $V_\phi$, target value network $V_{\bar{\phi}} \leftarrow V_\phi$, policy network $\pi_\psi$
\STATE Initialize learned temperature $\alpha$ and target-entropy schedule $\bar{\mathcal{H}}_t$ (cf. Remark~\ref{rk:SAC-1})
\STATE Initialize replay buffer $D \leftarrow \emptyset$ and fill it with $W$ warm-up transitions collected by $\pi_\psi$

\FOR{$t=1,\dots,\mathrm{budget}$}
    \STATE environment step: sample action $a \sim \pi_\psi(\cdot \mid s)$, next state $s' \sim p(\cdot \mid s,a)$
    \STATE Store transition $(s,a,r(s,a),d,s')$ in $D$, with $d$ the termination flag

    \IF{$t \bmod U = 0$}
        \STATE \textit{(Policy Evaluation Step)}
        \STATE Freeze a reference copy $V_{\phi_\mathrm{old}} \leftarrow V_\phi$ for value clipping
        \FOR{$k=1,\dots,K_V$}
            \STATE Sample minibatch $B = \{(s_i,a_i,r_i,d_i,s'_i)\}_{i=1}^b$ from $D$
            \STATE Compute soft target $T_i^V = r_i +  \alpha \mathcal{H}(\pi_\psi(\cdot|s_i)) + (1-d_i)\,\gamma V_{\bar{\phi}}(s_i')$, and the clipped value loss
            \begin{equation*}
                 L_\mathrm{V}(\phi) = \frac{1}{b} \sum_{i=1}^b  \max\Big\{ \big\{ V_\phi(s_i) - T_i^V\big\}^2,\, \big\{ V^\mathrm{clip}_\phi(s_i) - T_i^V\big\}^2 \Big\}\tag*{cf. Eq. \ref{eq::LV}}
            \end{equation*}
            \STATE with $V^\mathrm{clip}_\phi(s_i) = V_{\phi_\mathrm{old}}(s_i) + \mathrm{clip}\big(V_\phi(s_i) - V_{\phi_\mathrm{old}}(s_i), -\epsilon, \epsilon\big)$
            \STATE update the value network via gradient descent on $\nabla_\phi L_\mathrm{V}$
            \STATE \textit{(Target network update)} $\bar{\phi} \leftarrow \tau \phi + (1-\tau)\bar{\phi}$
        \ENDFOR
        \STATE
        \FOR{$k=1,\dots,K_A$}
            \STATE Sample minibatch $B = \{(s_i,a_i,r_i,d_i,s'_i)\}_{i=1}^b$ from $D$
            \STATE Compute advantage targets $T_i^A = r_i + \alpha \mathcal{H}(\pi_\psi(\cdot|s_i)) + (1-d_i)\,\gamma V_\phi(s_i') - V_\phi(s_i)$, and the advantage loss
            \begin{equation*}
              L^\mathrm{A}_\mathrm{critic}(\theta) = \frac{1}{b} \sum_{i=1}^b \left\{A_\theta(s_i,a_i) - T_i^A\right\}^2 \tag*{cf. Eq. \ref{eq:LAalpha}}
            \end{equation*}
            \STATE Update critic via gradient descent on $\nabla_\theta L^\mathrm{A}_\mathrm{critic}$
        \ENDFOR

        \STATE
        \STATE \textit{(Policy Improvement Step)}
        \FOR{$k=1,\dots,K_\pi$}
            \STATE Sample minibatch of states $\{s_i\}_{i=1}^b$ from $D$ and compute
            \begin{equation*}
                L_\pi(\psi) = \frac{1}{b} \sum_{i=1}^b \Big(
            \mathbb{E}_{a \sim \pi_\psi(\cdot \mid s_i)}
            \big[ -A_\theta(s_i,a) \big]
            - \alpha \mathcal{H}(\pi_\psi(\cdot \mid s_i)) \Big) \tag*{cf. Eq. \ref{eq:4}}
            \end{equation*}
            \STATE Update policy via gradient descent on $\nabla_\psi L_\pi$
        \ENDFOR

        \STATE
        \STATE \textit{(Temperature Step)} as in Algorithm~\ref{alg:SAC}
    \ENDIF
\ENDFOR
\end{algorithmic}
\end{algorithm}

\vspace{0.2cm}
\begin{remark}[Shared components]
\label{rk:ASAC-1}
    Data collection, the temperature step, the gradient-based updates and the hard target updates (now counted in value-network steps) are identical to SAC. By design, and in contrast to SAC, there is no critic ensemble: a single advantage network $A_\theta$ and a single value network $V_\phi$ are used.
\end{remark}

\vspace{0.1cm}
\begin{remark}[Consistency weight $\kappa$]
\label{rk:ASAC-4}
    The consistency term enforces $\mathbb{E}_{a\sim\pi_\psi}[A_\theta(s,a)] = 0$. In our experiments we set $\kappa = 0$, i.e.\ the advantage critic is trained on the regression loss alone.
\end{remark}

\subsection{\SACDA{}}

\begin{algorithm}[H]
\caption{SAC-DA}
\label{alg:DASAC}
\begin{algorithmic}
\STATE {\bfseries Initialize:} critic $A_\theta$, soft value network $V_\phi$, target networks $A_{\bar{\theta}} \leftarrow A_\theta$, $V_{\bar{\phi}} \leftarrow V_\phi$, policy network $\pi_\psi$
\STATE Initialize learned temperature $\alpha$ and target-entropy schedule $\bar{\mathcal{H}}_t$ (cf. Remark~\ref{rk:SAC-1})
\STATE Initialize replay buffer $D \leftarrow \emptyset$ and fill it with $W$ warm-up transitions collected by $\pi_\psi$

\FOR{$t=1,\dots,\mathrm{budget}$}
    \STATE environment step: sample action $a \sim \pi_\psi(\cdot \mid s)$, next state $s' \sim p(\cdot \mid s,a)$
    \STATE Store transition $(s,a,r(s,a),d,s')$ in $D$, with $d$ the termination flag

    \IF{$t \bmod U = 0$}
        \STATE \textit{(Policy Evaluation Step)}
        \FOR{$k=1,\dots,K_Q$}
            \STATE Sample minibatch $B = \{(s_i,a_i,r_i,d_i,s'_i)\}_{i=1}^b$ from $D$
            \STATE Combine advantages and values to Q-values
            \begin{equation*}
                  Q_{\theta,\phi}(s_i,a_i) = A_\theta(s_i,a_i) + V_\phi(s_i)- \alpha \mathcal{H}(\pi_\psi(\cdot|s_i)) - \kappa_\mathrm{d}\,\widehat{\IE}\bigl[A_\theta(s_i,\cdot)\bigr] \tag*{cf. Eq. \ref{eq::DAJointQ}}
            \end{equation*}
            \STATE Compute soft targets using the joint soft Q of the target networks
            \begin{equation*}
                T_i = r_i + (1-d_i)\, \gamma \, \Big( \mathbb{E}_{a' \sim \pi_\psi(\cdot \mid s'_i)} \big[ Q_{\bar{\theta}, \bar{\phi}}(s_i',a') \big] + \alpha \mathcal{H}(\pi_\psi(\cdot \mid s'_i)) \Big) \tag*{cf. Eq. \ref{eq:2}}
            \end{equation*}
            \STATE Calculate the critic loss
            \begin{equation*}
                L^\mathrm{DA}_\mathrm{critic}(\theta,\phi) = \frac{1}{b} \sum_{i=1}^b \left\{Q_{\theta,\phi}(s_i,a_i) - T_i\right\}^2 \tag*{cf. Eq. \ref{eq::DASACLoss}}
            \end{equation*}
            \STATE Update advantages and values jointly via gradient descent on $\nabla_{\theta,\phi} L^\mathrm{DA}_\mathrm{critic}$
            \STATE \textit{(Target network updates)} $\bar{\phi} \leftarrow \tau \phi + (1-\tau)\bar{\phi}$ and $\bar{\theta} \leftarrow \tau \theta + (1-\tau)\bar{\theta}$
        \ENDFOR
        \STATE
        \STATE \textit{(Policy Improvement Step)} as in Algorithm~\ref{alg:ASAC}
        \STATE \textit{(Temperature Step)} as in Algorithm~\ref{alg:SAC}
    \ENDIF
\ENDFOR
\end{algorithmic}
\end{algorithm}

\vspace{0.2cm}
\begin{remark}[Shared components]
\label{rk:DASAC-1}
    Data collection, the temperature step, the gradient-based updates and the hard target updates are identical to SAC. By design, a single advantage and value network are used, as in \SACA{}. In contrast to \SACA{} $A_\theta$ and $V_\phi$ are trained jointly on a single loss. Both target networks $A_{\bar{\theta}}$ and $V_{\bar{\phi}}$ receive a hard update every $N$ joint critic steps.
\end{remark}

\begin{remark}[Regularizer weight $\kappa_\mathrm{d}$]
\label{rk:DASAC-4}
  $\kappa_\mathrm{d} = 0$ worked best in preliminary experiments and is used throughout (Section~\ref{sec:advantages}). The split into $A_\theta$ and $V_\phi$ is then not identifiable, and \SACDA{} tests the dueling parametrization of the soft Q-critic.
\end{remark}

\subsection{SAC-\texorpdfstring{$\lambda$}{lambda}}

\begin{algorithm}[H]
\caption{SAC-$\lambda$}
\label{alg:PSAC}
\begin{algorithmic}
\STATE {\bfseries Initialize:} critics $Q_{\theta_1}$, $Q_{\theta_2}$, target critics $Q_{\bar{\theta}_j} \leftarrow Q_{\theta_j}$, policy network $\pi_\psi$
\STATE Initialize learned temperature $\alpha$ and target-entropy schedule $\bar{\mathcal{H}}_t$ (cf. Remark~\ref{rk:SAC-1})

\FOR{$t=1,\dots,\mathrm{budget}$}
    \STATE environment step: continue $E$ parallel environments for $T$ steps under $\pi_\psi$, giving trajectory segments $\{\tau_i\}_{i=1}^E$, $\tau_i = (s_{i,n},a_{i,n},r_{i,n},d_{i,n},e_{i,n},s'_{i,n})_{n=1}^T$, with $d$ the termination and $e$ the termination-or-truncation flag
    \STATE
    \STATE \textit{(Calculate $\lambda$-returns)}
    \STATE Compute soft values $\bar{V}_{i,n} = \mathbb{E}_{a' \sim \pi_\psi(\cdot \mid s'_{i,n})}\big[\operatorname*{agg}_{j=1,2} Q_{\bar{\theta}_j}(s'_{i,n},a') - \alpha \log(\pi_\psi(a' \mid s'_{i,n}))\big]$
    \FOR{$i=1,\dots,E$}
        \STATE set $R_{i,T} = r_{i,T} + \gamma\,(1-d_{i,T})\,\bar{V}_{i,T}$
        \FOR{$n=T-1,\dots,1$}
            \STATE $\lambda_{i,n} = \lambda\,(1-e_{i,n})$
            \STATE $R_{i,n} = r_{i,n} + \gamma\,\big\{(1-\lambda_{i,n})(1-d_{i,n})\,\bar{V}_{i,n} + \lambda_{i,n} \big( R_{i,n+1} + \alpha\mathcal{H}(\pi_\psi(\cdot \mid s'_{i,n})) \big) \big\}$
            \COMMENT{cf.\ Eq. \ref{eq::n-step Returns recursive}}
        \ENDFOR
    \ENDFOR
    \STATE
    \STATE \textit{(Policy Evaluation Step)}
    \FOR{$k=1,\dots,K_Q^\mathrm{ep}$ epochs, and each of $K_Q^\mathrm{mb}$ shuffled minibatches $B \subset \{(i,n)\}$}
        \STATE For each critic calculate value loss
        \begin{equation*}
             L_\mathrm{critic}^{\lambda}(\theta_j) = \frac{1}{|B|} \sum_{(i,n) \in B} \left\{ Q_{\theta_j}(s_{i,n}, a_{i,n}) - R_{i,n}\right\}^2\tag*{cf. Eq. \ref{eq::LossPSAC}}
        \end{equation*}
        \STATE Update the critics via gradient descent on $\nabla_{\theta_j} L_\mathrm{critic}^{\lambda}$
        \STATE \textit{(Target network update)} $\bar{\theta}_j \leftarrow \tau \theta_j + (1-\tau)\bar{\theta}_j$
    \ENDFOR
    \STATE
    \STATE \textit{(Policy Improvement Step)}
    \STATE Compute
    \begin{equation*}
        L_\pi(\psi) = \frac{1}{E} \sum_{i=1}^E \frac{1}{T} \sum_{n=1}^{T} \mathbb{E}_{a \sim \pi_\psi(\cdot \mid s_{i,n})} \big[ -\operatorname*{agg}_{j=1,2} Q_{\theta_j}(s_{i,n},a) + \alpha\,\log(\pi_\psi(a \mid s_{i,n})) \big] \tag*{cf. Eq. \ref{eq:4}}
    \end{equation*}
    \STATE Update policy via gradient descent on $\nabla_\psi L_\pi$
    \STATE
    \STATE \textit{(Temperature Step)} as in Algorithm~\ref{alg:SAC}, on the states of the current rollout
\ENDFOR
\end{algorithmic}
\end{algorithm}

\vspace{0.2cm}
\begin{remark}[Shared components]
\label{rk:PSAC-1}
    The critic ensemble and its aggregation $\operatorname*{agg}$, the discrete/continuous handling, the temperature step, the optimizer and the hard target updates (counted in critic gradient steps) are identical to SAC. 
\end{remark}

\vspace{0.1cm}
\begin{remark}[On-policy training]
\label{rk:PSAC-2}
    SAC-$\lambda$ uses no replay buffer. Every iteration collects $E \cdot T$ fresh transitions, where the segments continue from the previous iteration rather than starting new episodes. The $\lambda$-returns are computed once per iteration with the current target critics and temperature, and they stay fixed during the subsequent updates. The critics are trained for $K_Q^\mathrm{ep}$ epochs over $K_Q^\mathrm{mb}$ shuffled minibatches each, while the policy and the temperature each take a single full-batch gradient step per iteration.
\end{remark}

\begin{remark}[Soft Peng's Q($\lambda$)]
\label{rk:PSAC-3}
  With $\bar{V}(s) = \IE_{a\sim\pi_\psi}[Q_{\bar\theta}(s,a)] + \alpha\mathcal{H}(\pi_\psi(\cdot|s))$, the recursion equals Eq.~\ref{eq::n-step Returns recursive}: with weight $1-\lambda$ the return bootstraps from the soft value of the next state under the current policy rather than from the action actually taken, and with weight $\lambda$ it continues with the next return plus the entropy bonus at the next state. This is the soft analogue of Peng's Q($\lambda$)~\citep{peng1994incremental}; unlike \citet{kozuno2021revisiting}, we keep the entropy bonus at every step. At segment ends and episode boundaries ($e_{i,n}=1$) the trace is cut, and bootstrapping is suppressed only on termination ($d_{i,n}=1$), so on truncation the return bootstraps from the true final state.
\end{remark}

\subsection{\SACG{}}

In order to get a suitable gradient estimator for our empirical policy loss
\begin{equation*}
    L_\pi(\psi) = \frac{1}{b} \sum_{i=1}^b \IE_{a\sim \pi_\psi(\cdot|s_i)}[-\hat{A}(s_i,a) + \alpha \log(\pi_\psi(a|s_i))],
\end{equation*}
some care is needed. We only have access to the advantage estimator $\hat{A}(s_i,a_i)$ at the sampled action $a_i$, not under all actions, so the loss above cannot be implemented directly. The soft policy gradient theorem \citep{shiSoft2019, choeMaxi} resolves this, and the same proof yields
\begin{align*}
    \nabla_\psi \IE_{a\sim \pi_\psi(\cdot|s)}[\hat{A}(s,a) - \alpha \log(\pi_\psi(a|s))] = \IE_{a\sim \pi_\psi(\cdot|s)}[\nabla_\psi \log(\pi_\psi(a|s)) \{\hat{A}(s,a) - \alpha (\log(\pi_{\mathrm{sg}(\psi)}(a|s)) + 1)\}],
\end{align*}
where $\mathrm{sg}(\cdot)$ refers to a stop gradient. We can use this expression to obtain a gradient estimator for the above loss as
\begin{equation}
    \label{eq:helper2}
    \nabla_\psi L_\pi(\psi) = \frac{1}{b} \sum_{i=1}^b \nabla_\psi \log(\pi_\psi(a_i|s_i)) \{-\hat{A}(s_i,a_i) + \alpha (\log(\pi_{\mathrm{sg}(\psi)}(a_i|s_i)) + 1)\}
\end{equation}

\begin{algorithm}[h]
\caption{SAC-GAE}
\label{alg:GAESAC}
\begin{algorithmic}
\STATE {\bfseries Initialize:} soft value network $V_\phi$, policy network $\pi_\psi$
\STATE Initialize learned temperature $\alpha$ and target-entropy schedule $\bar{\mathcal{H}}_t$ (cf. Remark~\ref{rk:SAC-1})

\FOR{$t=1,\dots,\mathrm{budget}$}
    \STATE environment step: continue $E$ parallel environments for $T$ steps under $\pi_\psi$
    \STATE Store trajectory segments in trajectory batch $\{\tau_i\}_{i=1}^E$, $\tau_i = (s_{i,n},a_{i,n},r_{i,n},d_{i,n},e_{i,n},s'_{i,n})_{n=1}^T$, with $d$ the termination and $e$ the termination-or-truncation flag
    \STATE Freeze a reference copy $V_{\phi_\mathrm{old}} \leftarrow V_\phi$
    \STATE
    \STATE \textit{(Calculate soft GAEs)}
    \COMMENT{cf.\ Eq. \ref{eq:SoftGAE}}
    \FOR{$i=1,\dots,E$}
        \STATE set $A_{i,T+1} = 0$
        \FOR{$n=T,\dots,1$}
            \STATE $\delta_{i,n} = r_{i,n} + \alpha\,\mathcal{H}(\pi_\psi(\cdot|s_{i,n})) + \gamma\,(1-d_{i,n})\, V_{\phi_\mathrm{old}}(s'_{i,n}) - V_{\phi_\mathrm{old}}(s_{i,n})$
            \STATE $A_{i,n} = \delta_{i,n} + \gamma\,\lambda\,(1-e_{i,n})\,A_{i,n+1}$
            \STATE $R_{i,n} = A_{i,n} + V_{\phi_\mathrm{old}}(s_{i,n})$
        \ENDFOR
    \ENDFOR

    \STATE
    \STATE \textit{(Policy Evaluation Step)}
    \FOR{$k=1,\dots,K_V^\mathrm{ep}$ epochs, and each of $K_V^\mathrm{mb}$ shuffled minibatches $B \subset \{(i,n)\}$}
        \STATE Calculate clipped value loss
        \begin{equation*}
             L_\mathrm{V}(\phi) = \frac{1}{2|B|} \sum_{(i,n) \in B} \max\Big\{ \left\{ V_\phi(s_{i,n}) - R_{i,n}\right\}^2,\, \left\{ V^\mathrm{clip}_\phi(s_{i,n}) - R_{i,n}\right\}^2 \Big\}\tag*{cf. Eq. \ref{eq::LV}}
        \end{equation*}
        \STATE with $V^\mathrm{clip}_\phi(s) = V_{\phi_\mathrm{old}}(s) + \mathrm{clip}\big(V_\phi(s) - V_{\phi_\mathrm{old}}(s), -\epsilon, \epsilon\big)$
        \STATE Update the value network via gradient descent on $\nabla_\phi L_\mathrm{V}$
    \ENDFOR

    \STATE
    \STATE \textit{(Policy Improvement Step)}
    \STATE Normalize advantages $\tilde{A}_{i,n} = (A_{i,n} - \mu_A)/(\sigma_A + 10^{-8})$ over the batch and compute

    \begin{equation*}
        L_\pi(\psi) = \frac{1}{E} \sum_{i=1}^E \frac{1}{T} \sum_{n=1}^{T} \log(\pi_\psi(a_{i,n}|s_{i,n})) \{- \tilde{A}_{i,n} + \alpha (\log(\pi_{\mathrm{sg}(\psi)}(a_{i,n}|s_{i,n})) + 1)\} \tag*{cf. Eq. \ref{eq:helper2}}
    \end{equation*}
    \STATE Update policy via gradient descent on $\nabla_\psi L_\pi$

    \STATE
    \STATE \textit{(Temperature Step)} as in Algorithm~\ref{alg:SAC}, on the states of the current rollout
\ENDFOR
\end{algorithmic}
\end{algorithm}

\vspace{0.2cm}
\begin{remark}[No target network]
\label{rk:GAESAC-2}
    In contrast to all other variants, SAC-GAE uses no target network. The advantages and returns are computed once per iteration with the value network $V_{\phi_\mathrm{old}}$ from before the update and stay fixed during the subsequent updates. The same $V_{\phi_\mathrm{old}}$ serves as reference for the value clipping with $\epsilon = 10$.
\end{remark}

\subsection{Replay-buffer variants}
\label{sec:buffer-variants}

\SACLR{} and \SACGR{} combine the off-policy training loop of SAC (Algorithm~\ref{alg:SAC}) with the return estimators of SAC-$\lambda$ (Algorithm~\ref{alg:PSAC}) and SAC-GAE (Algorithm~\ref{alg:GAESAC}), respectively. Data collection, the replay buffer, the update schedule ($K_Q$ critic steps, then $K_\pi$ policy steps and $K_\alpha$ temperature steps every $U$ environment steps), the temperature step and the hard target updates are exactly as in SAC. The networks and losses are those of the on-policy counterparts: \SACLR{} uses the twin critics $Q_{\theta_j}$ and the SAC policy loss, and \SACGR{} uses a single value network $V_\phi$ without target network, the untanh'd Gaussian policy and the policy-gradient loss of Eq.~\eqref{eq:helper2}. The only structural change is how a minibatch is formed: instead of $b$ independent transitions, each gradient step samples contiguous traces from $D$ and computes the returns on them with the current networks.

\vspace{0.1cm}
\noindent The critic (resp.\ value) loss is then the one of the on-policy counterpart on $B$, i.e.\ $\frac{1}{b}\sum_{(i,m)\in B}\{Q_{\theta_j}(s_{i,m},a_{i,m}) - R_{i,m}\}^2$ for each critic in \SACLR{}, and the value loss of Algorithm~\ref{alg:GAESAC} in \SACGR{}. The policy loss of \SACGR{} uses the advantages of a separately sampled trace minibatch, computed after the critic updates and normalized over that minibatch. \SACLR{} and the temperature step only need the states $s_{i,m}$ and use no returns.

In contrast to the on-policy variants, where returns are computed once per rollout and kept fixed, here they are recomputed for every sampled minibatch with the current target critics (\SACLR{}) or the current value network (\SACGR{}), and with the current policy and temperature. Hence the traces are stale, but their returns are not.

\end{document}

%% file: _util/commands.tex
\newcommand{\R}{\ensuremath{\mathbb{R}}}   

\newcommand{\IE}{\ensuremath{\mathbb{E}}}

\def\A{{\cal A}}
\def\S{{\cal S}}

\theoremstyle{plain}
\newtheorem{theorem}{Theorem}[section]

\newtheorem{lemma}[theorem]{Lemma}

\theoremstyle{definition}

\theoremstyle{remark}
\newtheorem{remark}[theorem]{Remark}